\documentclass[a4paper,fleqn]{cas-sc}

\usepackage[authoryear, sort]{natbib}
\usepackage{textcomp}
\usepackage{multicol,multicap,graphicx}
\usepackage{subfig}
\usepackage{xcolor}

\def\tsc#1{\csdef{#1}{\textsc{\lowercase{#1}}\xspace}}
\tsc{WGM}
\tsc{QE}

\begin{document}
\let\WriteBookmarks\relax
\def\floatpagepagefraction{1}
\def\textpagefraction{.001}



\title [mode = title]{A General Mobile Manipulator Automation Framework for Flexible Manufacturing in Hostile Industrial Environments}  



%

\author[1]{Can Pu$^\dagger$}[style=chinese]



\ead{can.pu@amigaga.com}


\credit{Conceptualization, Methodology, Experiment, Writing - original draft, Funding acquisition}

\affiliation[1]{organization={Shenzhen Amigaga Technology Co. Ltd.},
            addressline={Building B, U+ Research Center, Gushu 1st Road, Baoan District}, 
            city={Shenzhen},
          citysep={}, 
            postcode={518000}, 
            country={China}}

\author[1]{Chuanyu Yang$^\dagger$}[style=chinese]

\cormark[1]


\credit{Methodology, Experiment, Review \& editing, Funding acquisition.}

\author[1,2]{Jinnian Pu}[]

\ead{jipu0216@uni.sydney.edu.au}

\credit{Experiment data analysis and visualization}

\affiliation[2]{organization={The University of Sydney},
	addressline={City Road}, 
	city={Camperdown/Darlington},
	citysep={}, 
	postcode={NSW 2006}, 
	country={Australia}}

\author[3]{Robert B. Fisher}[]

\ead{rbf@inf.ed.ac.uk}

\credit{Review \& editing, Supervision}

\affiliation[3]{organization={School of Informatics, University of Edinburgh},
            addressline={1.11 Bayes Centre, 47 Potterrow}, 
            city={Edinburgh},
          citysep={}, 
            postcode={EH89BT}, 
            country={UK}}

\cortext[1]{ Corresponding author at: Shenzhen Amigaga Technology Co. Ltd.  Email address: chuanyu.yang@amigaga.com (C. Yang) \newline \indent \indent $\dagger$ denotes: equally contributed}


\begin{abstract}[SUMMARY]
\textcolor{black}{To enable a mobile manipulator to perform human tasks from a single teaching demonstration is vital to flexible manufacturing}. We call our proposed method MMPA (Mobile Manipulator Process Automation with One-shot Teaching). Currently, there is no effective and robust MMPA framework which is not influenced by harsh industrial environments and the mobile base's parking precision. The proposed MMPA framework consists of two stages: collecting data (mobile base's location, environment information, end-effector's path) in the teaching stage for robot learning;  letting the end-effector repeat the nearly same path as the reference path in the world frame to reproduce the work in the automation stage. More specifically, in the automation stage, the robot navigates to the specified location without the need of a precise parking. Then, based on colored point cloud registration, the proposed IPE (Iterative Pose Estimation by Eye \& Hand) algorithm could estimate the accurate 6D relative parking pose of the robot arm base without the need of any marker. Finally, the robot could learn the error compensation from the parking pose's bias to modify the end-effector's path to make it repeat a nearly same path in the world coordinate system as recorded in the teaching stage. Hundreds of trials have been conducted with a real mobile manipulator to show the superior robustness of the system and the accuracy of the process automation regardless of the harsh industrial conditions and parking precision. For the released code, please contact \href{marketing@amigaga.com}{marketing@amigaga.com}
\end{abstract}


\maketitle

\section{Introduction}\label{introduction} Mobile manipulators (def: a mobile platform \textcolor{black}{carrying} a robot arm) have attracted much interest in universities and industry because of \textcolor{black}{the combination of} flexible locomotion and dexterous manipulation~\citep{survey_application_mobile_manipulator_2022,survey_mobile_manipulator_2019}. A mobile manipulator is able to \textcolor{black}{undertake} multiple different industrial tasks in a large workspace (rather than at a fixed work station), and can be widely used in many flexible manufacturing tasks, such as quality inspection in factories, workpiece loading and unloading in workshops, and painting. It is cumbersome and time-consuming to reprogram for each robot application. Thus, it is vital for flexible manufacturing to minimize the human effort spent on teaching the robot, and \textcolor{black}{to} ensure the mobile manipulator is able to automate and redo multiple desired industrial tasks in a large workspace flexibly with \textcolor{black}{one-shot teaching} from a human operator. This one-shot teaching and automation pipeline of a mobile manipulator for flexible manufacturing is referred to as Mobile Manipulator Process Automation with One-shot Teaching (MMPA)\footnote{\textcolor{black}{All the abbreviations in this manuscript are listed in Appendix~\ref{Abbreviations}. List of Abbreviations.}}.

\begin{figure}
	\begin{center}
		\subfloat{\includegraphics[width=1\linewidth]{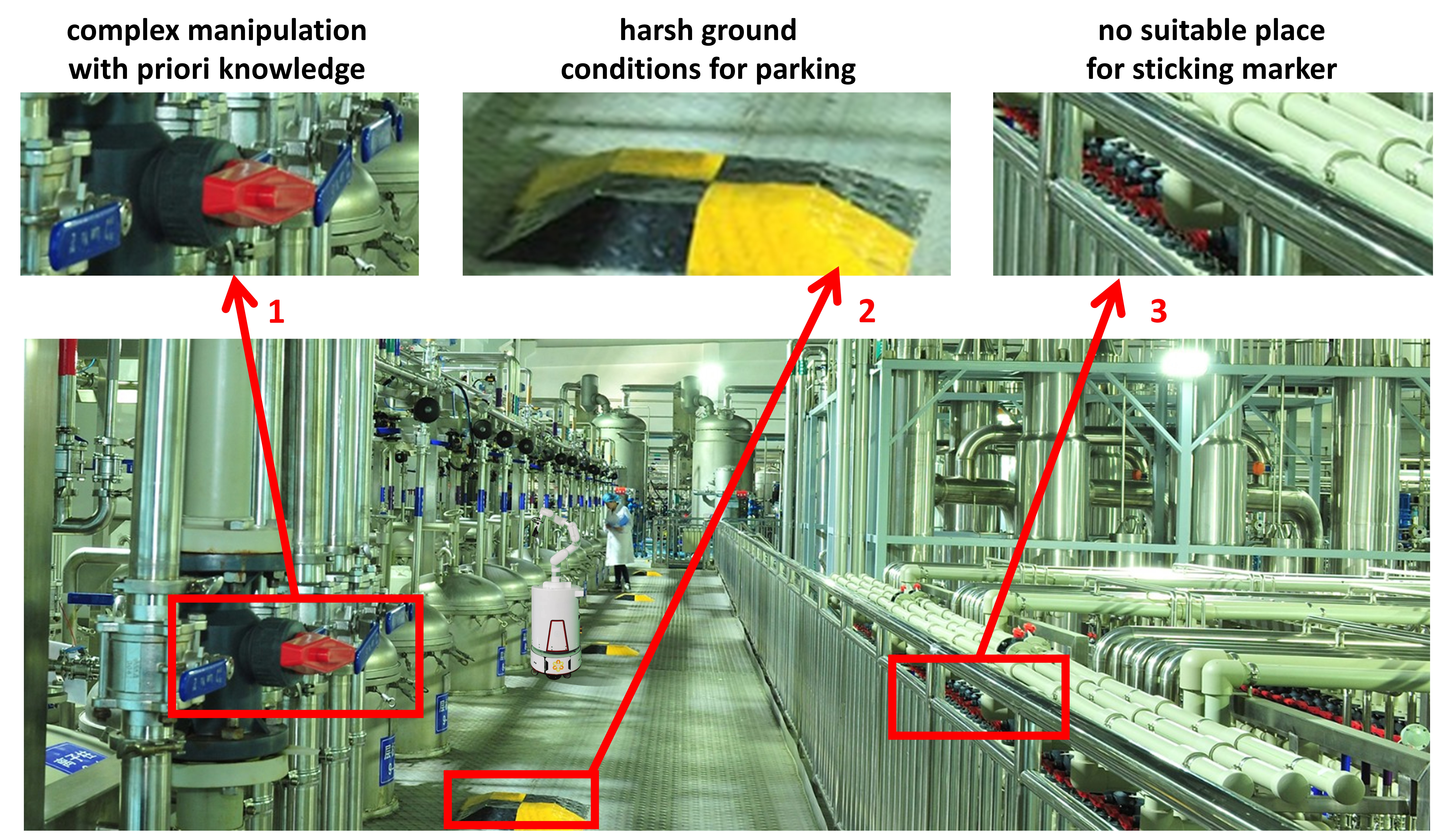}}	
	\end{center}
	\caption[scene-and-application]{A typical workshop example with harsh industrial conditions and our mobile manipulator. 1 The mobile manipulator is required to perform complex manipulation in 3D space with priori knowledge such as how to switch the valve rather than simple workpiece loading and unloading. 2 The ground consists of not only 2D plane but also 3D speed bump where a working robot parks on. 3 The guard bars and pipes are not suitable to stick any marker for error compensation.}
	\label{fig:scene}
\end{figure}

 Current mobile manipulator process automation approaches consist of two stages: in the first stage the mobile base's location and the robot arm's path are recorded; in the second stage the mobile manipulator will park at the same location and replay the arm's path as recorded in the first stage. Such approaches rely heavily on precise parking of the robot base and are easily affected by harsh industrial conditions. As the robot arm's path is replayed in the exact same manner as recorded in the teaching stage, a minor error of the parking \textcolor{black}{position} will lead to an incorrect robot arm manipulation.
However, parking precisely is hard to achieve in a real harsh industrial environment. For example, a rough, cratered, slippery ground will affect robot control accuracy\footnote{ Note: on the market, a mobile base with high control accuracy (millimeter level) is 10+ times more expensive than a regular mobile base (centimeter level) we use.} definitely~\citep{terrain-parameter1-2018,terrain-parameter2-2021,fuzzy-control-2021,impedance-control-2022}. Additionally, the localization accuracy from SLAM (\textcolor{black}{Simultaneous Localization And Mapping})~\citep{preintegration-gnss-imu-in-vehicle-2021, calibration-IMU-Odometer-2022,gmapping2007, cartographer2016, slam-survey1-2021, multi-sensor-fusion-2022,slam-survey2-2021} is not high enough for precise parking in MMPA. Matching the 2D template marker's images from the teaching and automation stages~\citep{iterative-learning-error-compensation2021} could provide a relative 3D pose for parking error compensation but can  only be used to move the robot base on a 2D plane rather than a 3D surface. Lastly, there is no existing 6D pose recovery algorithm~\citep{survey-point-cloud-matching-2021,GO-ICP2015,ICP-1-2022,ICP-2-2022,colored-point-cloud-registration-2017,3DMatch-2017, point-cloud-matching-descriptor-2021,point-cloud-matching-descriptor2-2021,fast-global-registration2016, CPD, CPD2,DUGMA, GMM-point-cloud-matching} that provides an accurate and robust 6D pose estimation (\textcolor{black}{The 6D pose is [translation on $x$ axis, translation on $y$ axis, translation on $z$ axis, roll, pitch, yaw]}) for parking pose error compensation to meet the requirement of MMPA.  As described above, Figure~\ref{fig:scene} shows a real typical workshop example with a harsh industrial environment where our developed robot is working. \textcolor{black}{The two articles~\citep{robotics_design_for_harsh_environment1,robotics_design_for_harsh_environment2} gives a good overview of the design of the autonomous robotics system against the harsh environment in the real-life setting. }   

The proposed MMPA method is capable of dealing with the problem of bad ground conditions with a novel iterative pose estimation approach to compensate for the error introduced by imprecise parking. The proposed MMPA framework has two stages, a one-shot teaching stage and an automation stage. In the one-shot teaching stage, a human teaches the mobile manipulator to perform a specific tasks. Information such as the location of the robot base, working environment information and the path of the robot arm's end-effector are collected, to allow for the reproduction of the task in the automation stage. In the automation stage, the robot first navigates to the specified location without the need of precise parking. Then, based on the global colored point cloud registration, the proposed IPE (Iterative Pose Estimation by Eye \& Hand) algorithm provides an accurate relative 6D parking pose without the need of any marker. Finally, the mobile manipulator uses the relative 6D parking pose to calibrate and adjust the path of the robot arm's end-effector for performing complex tasks taught by the human operator during the one-shot learning stage. In this work, hundreds of trials have been conducted with a real mobile manipulator to show that our proposed MMPA framework and IPE algorithm are capable of calibrating and adjusting the path of the robot arm end-effector to compensate for the parking error and guarantee the robustness of the system and the accuracy of the process automation without worrying about ground conditions and parking precision.

\textcolor{black}{
	The proposed MMPA could bring much benefit to robot developers and customers who need the joint automation of flexible locomotion and dexterous manipulation. Currently, robot developers need to spend much time in programming robots for various tasks, which limits the robot application development efficiency and increases the development fee drastically. With the proposed MMPA, robot developers could develop simple robot automation tasks with human demonstration, which is also important to the customers without professional knowledge of robotics. The customers could adapt the mobile manipulator flexibly to a new task by themselves without the need of professional on-the-spot service, which reduces the maintenance fee and waiting time. With the proposed MMPA technique, one mobile manipulator could replace at least three workers\footnote{\textcolor{black}{The robot could work 24 hours a day while a human worker only works for 8 hours a day}} day and night, which decreases the labor cost. Additionally, compared to human workers, robot workers do not require human resource management and can work in hazardous environments, reducing the management burden for managers. Currently the proposed MMPA technique could be used to develop many real applications in various areas (e.g.: material handling and logistics~\citep{AMR}). The mobile manipulator could be used to load or unload the workpiece or electronical components and convey them to different designated places in some semi-conductor factories or mechanical companies. In an e-commerce warehouse, the mobile manipulator could be used to pick different goods on the shelves, pack them and deliver them to the postmen. In a hospital, the mobile manipulator could be used to take out the harmful medical waste and convey it to an appointed place without any intervention of human. Much more real applications are being developed on the way.   	
}

The remainder of this paper is structured as follows. Section~\ref{related-works} presents previous research about MMPA's progress, MMPA's impact factors, and 6D pose recovery algorithms. Section~\ref{methodology} presents the proposed algorithm for mobile manipulator process automation by one-shot teaching. Section~\ref{Experiment} demonstrates the robustness of the system and assesses the accuracy of the process automation regardless of the ground conditions and parking precision based on hundreds of trials using the real mobile manipulator. Section~\ref{Conclusion} presents a summary of the work.

\textbf{Contributions in this paper:}  

(1)   An accurate 6D pose estimation algorithm: IPE (Iterative Pose Estimation by Eye \& Hand);

(2)	An adaptive online path learning method to calibrate the end-effector's trial in 3D space effectively; 

(3)	An effective and robust MMPA framework that is able to compensate for the errors caused by ground conditions and low parking precision;

\section{Related Works}\label{related-works}
Mobile manipulators combine the advantages of both the mobile base's mobility~\citep{AMR} and the articulated arm's dexterity, which makes it an ideal tool to perform multiple manipulation tasks that cover different locations in a large workspace. Research \textcolor{black}{into} mobile manipulators has a long history, dating back to \textcolor{black}{the last century}~\citep{mobile-manipulator1996,mobile-manipulator1999, mobile-manipulator1992}. \textcolor{black}{The most recent surveys~\citep{survey_application_mobile_manipulator_2022,survey_mobile_manipulator_2019} give an explicit progress overview of the mobile manipulator's hardware system, software system and the new applications.} One of the interesting advanced technique is mobile manipulator process automation \textcolor{black}{because it requires the robot to autonomously reproduce the work taught by humans.} Currently, the main ideas to realize mobile manipulator process automation in a large workspace \textcolor{black}{are}: (1)  In the teaching stage, record the mobile base's location data and the arm's manipulation path data into the database. (2) In automation stage, let the mobile base navigate and park at the location from the teaching stage with high accuracy (e.g.: position accuracy on the $x$, $y$ axis < 0.2 $cm$, orientation accuracy < $0.1^{\circ}$) first; then let the robot arm replay the previously recorded manipulation path data from the database to reproduce the work that humans taught. Given the actuation accuracy of most economic collaborative robot arms is not above 0.1 $mm$, parking the mobile platform extremely accurately is the key for mobile manipulator process automation.

To make the robot base park precisely, the researchers have to deal with many aspects within the robot system, such as: sensor calibration~\citep{calibration-odometer-2018,calibration-IMU-Odometer-2022,preintegration-gnss-imu-in-vehicle-2021,sensor-fusion-2019}, localization~\citep{slam-survey1-2021,slam-survey2-2021,gmapping2007, cartographer2016, multi-sensor-fusion-2022, calibration-IMU-Odometer-2022,preintegration-gnss-imu-in-vehicle-2021}, control accuracy~\citep{terrain-parameter1-2018,terrain-parameter2-2021,fuzzy-control-2021,impedance-control-2022}, error compensation~\citep{robust-learning-control-2019,iterative-learning-error-compensation2021}. Single sensor calibration could rectify the output value by using a noise model to make it more accurate. For example, \textcolor{black}{the researchers~\citep{calibration-odometer-2018}} improve the accuracy of the odometer (which can provide the velocity or mileage of a vehicle) through use of Coriolis effects from three perspectives (scale factor, misalignment and level arm with inertial measurement unit). \textcolor{black}{Multiple sensor calibration~\citep{sensor-fusion-2019,preintegration-gnss-imu-in-vehicle-2021, calibration-IMU-Odometer-2022}} provides the relative pose among multiple different sensors for sensor fusion to improve the accuracy, such as using preintegration theory for IMU (\textcolor{black}{Inertial Measurement Unit}) and odometer self-calibration~\citep{calibration-IMU-Odometer-2022,preintegration-gnss-imu-in-vehicle-2021}, calibrating gyroscope and magnetometer for data fusion~\citep{sensor-fusion-2019}. However, sensor calibration could slightly reduce the error given the sensor's inherent properties or mounting positions. \textcolor{black}{Thus, it contributes little to precise parking.} In order to park accurately, mobile base localization is important. Currently there are many popular SLAM algorithms, especially, the Lidar-based SLAM~\citep{gmapping2007, cartographer2016}. \textcolor{black}{ The main feauture of the Lidar-based SLAM~\citep{slam-survey1-2021,slam-survey2-2021} is that it uses Lidar scanner to input the position data for mapping and localization. The Lidar-based SLAM consists of two main categories: 2D Lidar SLAM (e.g.: Gmapping~\citep{gmapping2007}, Cartographer~\citep{cartographer2016}) to generate a 2D map for localization (usually) indoors and 3D Lidar SLAM (e.g.: Lego-loam~\citep{Lego-loam}) to generate a 3D map for localization (usually) outdoors.
} In order to get a more robust and accurate localization, it is a trend to fuse data from multiple sensors (e.g.: lidar, odometer, imu, etc.) using Kalman Filter~\citep{Kalman-Filter-1995,Kalman-Filter-in-robotics-2021}, Graph-based methods~\citep{multi-sensor-fusion-2022, calibration-IMU-Odometer-2022,preintegration-gnss-imu-in-vehicle-2021} etc. . Nevertheless, their final localization accuracy is still limited to centimetre-level, which is still too large to meet the requirements of mobile manipulator process automation. Besides the above sensor calibration and localization, the mobile platform's control accuracy in the workspace is vital as well. Much research~\citep{terrain-parameter1-2018,terrain-parameter2-2021,fuzzy-control-2021,impedance-control-2022} about it has been done in recent years. For example, some researchers studied the terrain property~\citep{terrain-parameter1-2018,terrain-parameter2-2021} for the robot's control to handle the robot's slide etc. Some researchers proposed different improved control methods, such as fuzzy control~\citep{fuzzy-control-2021}, impedance control~\citep{impedance-control-2022}, etc. However, it is challenging to model the errors of the driving mobile base on different unknown terrain in a uniform way to ensure a robust and accurate control accuracy. Thus, parking precisely is hard in real-world scenarios where the environmental situation is unknown, which limits the mobile manipulator process automation's reliability.

Even if the mobile manipulator drives on a well known plane with high control accuracy, the localization accuracy still might not meet the requirements of the mobile manipulator process automation. In order to detect the localization's system error, iterative-learning error compensation methods~\citep{robust-learning-control-2019,iterative-learning-error-compensation2021} have been proposed recently. In this research~\citep{iterative-learning-error-compensation2021}, the researchers developed an extra eye-in-hand vision system to assist the mobile platform's localization. More specifically, they put an RGB camera at the end of the robot arm for perception and attached a QR (\textcolor{black}{Quick Response}) code on the workbench. In the teaching stage, the eye-in-hand vision system (the RGB camera on the arm) takes a photo of QR code using a fixed arm pose. In the automation stage, the mobile manipulator reaches the workbench and takes a photo of the previous QR code again using the previous same fixed arm pose. By matching those 2D QR code templates, the mobile base's relative 3D pose (position bias on the x, y axis and orientation angle) could be estimated. Then, the new relative 3D pose was used to rectify the mobile base's parking pose to make it park more precisely. Matching 2D QR code templates and rectifying the parking pose in an iterative way could make the mobile manipulator park precisely enough for a process automation. However, this method~\citep{iterative-learning-error-compensation2021} suffers from two big problems. Given the pose from 2D QR code template matching has only 3 degrees of freedom, the mobile manipulators could only be moved on a good 2D plane terrain rather than a cratered 3D surface. The second problem is that the condition of the 2D plane for robot base moving should be good enough for accurate robot base control in order to park precisely. 

To solve the two problems mentioned above~\citep{iterative-learning-error-compensation2021}, we use a \textcolor{black}{depth sensor (stereo vision camera)} on the arm to estimate the relative 6D pose of the mobile base by point cloud matching~\citep{survey-point-cloud-matching-2021}, which enables the estimation of the robot's pose, even on a rough 3D surface. By changing the robot arm end-effector's working path based on the relative 6D pose of the mobile base, our method does not require the robot base to park precisely, which eliminates the need of high control accuracy of the robot base and good quality ground condition. 

Currently, there are three categories of point cloud registration algorithms for estimating the relative 6D pose. \textcolor{black}{The recent survey~\citep{survey-point-cloud-matching-2021} introduces the taxonomy of the point cloud registration methods and their progress. A new cross-source point cloud benchmark~\citep{survey-point-cloud-matching-2021} is developed to evaluate the point cloud registration algorithms to solve cross-source challenges.} 
The first category of methods(e.g.:~\citep{GO-ICP2015,ICP-1-2022,ICP-2-2022,colored-point-cloud-registration-2017}) is based on ICP (\textcolor{black}{Iterative Closest Point})~\citep{ICP1992}, which estimates the rigid relative 3D pose between two point clouds in different views by minimizing the Euclidean distance between the corresponding points.  Exact point-to-point correspondences seldom exist, which leads to the low accuracy of the ICP-based methods. The second category is feature-based methods~\citep{3DMatch-2017, point-cloud-matching-descriptor-2021,point-cloud-matching-descriptor2-2021,fast-global-registration2016}, which extract the local descriptors from two point clouds first and then match them to recover the relative pose of the two point clouds. These methods are sensitive to strong noise and low density of the point clouds. That is, the noise and density of
the point cloud influences the local descriptors' extraction and can even cause the algorithm to crash if the noise is too strong or the density is too low. The third class uses probabilistic models~\citep{CPD, CPD2,DUGMA,GMM-point-cloud-matching} for point cloud registration. The probabilistic models are used to represent the structure of the point cloud, encoding the geometry distribution of the point cloud in 3D space. By calculating the maximum likelihood of the two probabilistic models, the relative pose can be estimated. This category of methods is more robust and accurate than the first and second category but its computation efficiency is low and limits many real-time applications.
Although those three categories of point cloud registration algorithm can estimate the relative pose from different views, accuracy and robustness decreases with the increase of the initial rotation and translation error between the two point clouds, which might not be able to meet the requirement for the high accuracy in the mobile manipulator process automation. In order to solve this problem, we propose a method called IPE:Iterative pose estimation by eye \& hand in Section~\ref{iterative-pose-estimation}.  

\textcolor{black}{
	Path planning, also known as motion planning, is an optimization problem that aims to find a valid collision-free trajectory to move the robot end-effector from a start point to an end point while satisfying certain constraints. Path planning problems are commonly solved using probabilistic sampling based algorithms such as rapidly exploring random tree (RRT*)~\citep{RRT2011}, probabilistic roadmap (PRM)~\citep{PRM1996}. Advance in path planning algorithms has allowed robotic arms to perform more complex and flexible tasks \citep{sucan2012openMotion, schulman2014motionPlanning}. Traditionally, the operation path of industrial robot arms within a factory is manually taught or programmed by a human with a dedicated teach pendant, which is tedious and time-consuming. In recent years, there has been an increase in interest on applying path planning algorithms to eliminate the time consuming manual teaching process. Path planning is typically used for manufacturing applications that involve processing industrial products with complex 3D geometry. Such applications include welding\citep{zhou2022onlineObstacle}, painting\citep{chen2020trajectoryPlanning}, grinding\citep{lv2020adaptiveTrajectory}, polishing\citep{mohsin2017roboticPolishing}, \textcolor{black}{cutting\citep{van2019accessibilityFor}, and inspection\citep{alexis2016aerialRobotic}. Coverage path planning is a type of path planning problem, which aims to determine a collision-free trajectory that uniformly passes through all points of an area or volume of interest while minimizing travel time, energy, and other costs\citep{galceran2013surveyOn}.}
} Manufacturing applications that require coverage over a specific area of interest include painting\citep{chen2020trajectoryPlanning}, grinding\citep{lv2020adaptiveTrajectory}, and inspection\citep{engemann2021robotAssisted}.

\section{Methodology}\label{methodology}
\textcolor{black}{The main goal of our proposed MMPA is to ensure that the robot arm's end-effector repeats the same path recorded during the teaching stage in the world coordinate system to reproduce the desired task even if the parking precision is low. \textbf{A simple practical demo video to show the pipeline (teaching stage and automation stage) is available: \url{https://youtu.be/n6D9T9GdqzI}.}
The whole pipeline consists of two stages: (1) data collection by one-shot teaching (Section \ref{one-shot-teaching}) to make the robot learn to perform a specific task; (2) replaying the newly-learned data by the mobile manipulator process automation (Section \ref{mobile-manipulator-process-automation}) to enable the robot to finish the task successfully and flexibly on its own.} 

\begin{figure}
	\centering
	\includegraphics[width=0.5\linewidth]{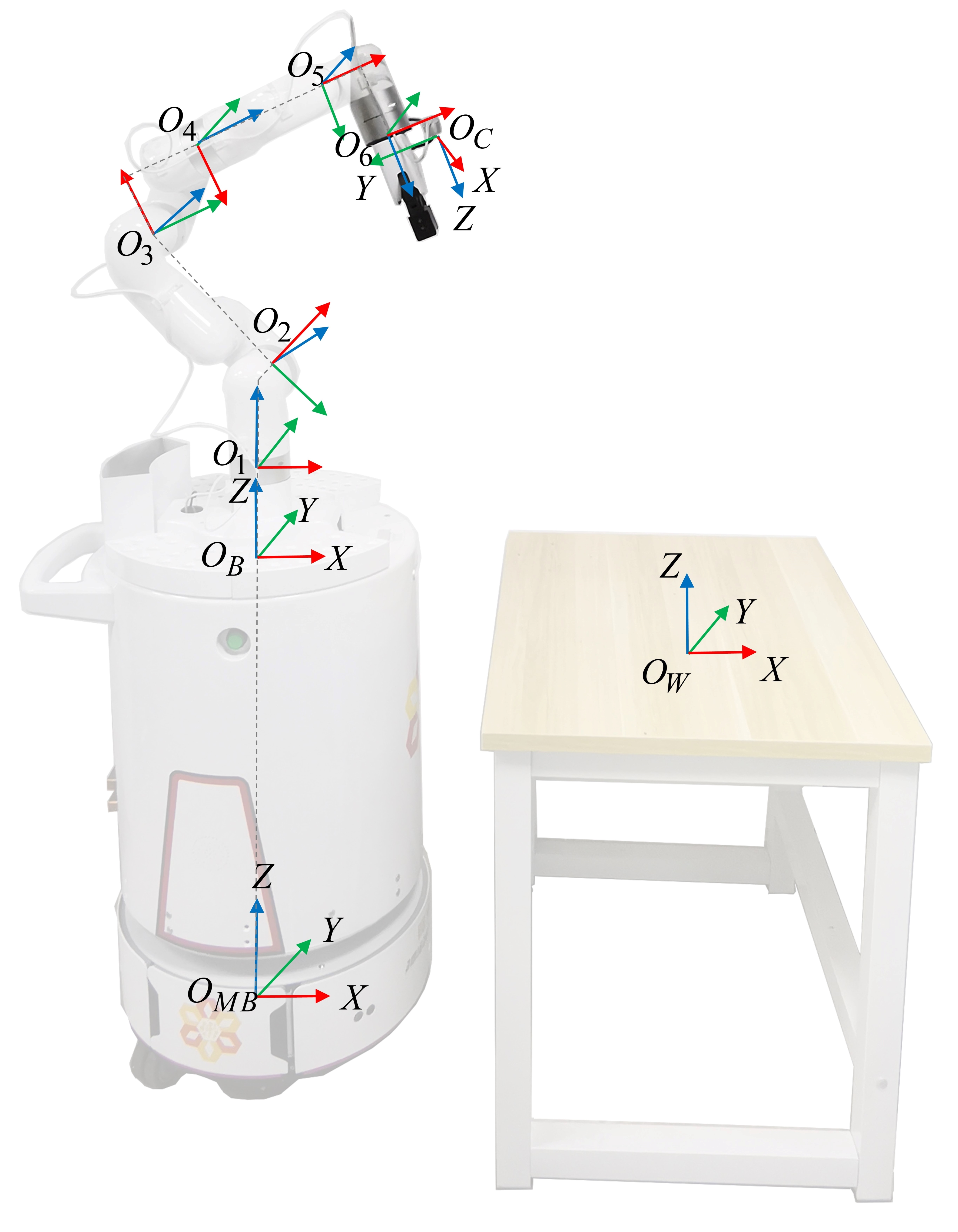}
	\caption{The figure shows the various coordinate frames in a mobile manipulator.}
	\label{fig:coordinate}
\end{figure}

Figure~\ref{fig:coordinate} shows different coordinate systems\footnote{\textcolor{black}{All the symbols in this manuscript are listed in Appendix~\ref{symbols}. List of Symbols.}}: mobile base coordinate system $XYZO_{MB}$, robot arm base coordinate system $XYZO_{B}$, depth camera coordinate system $XYZO_{C}$, world coordinate system $XYZO_{W}$, the joint coordinate system from $XYZO_{1}$ to $XYZO_{6}$.
 
\subsection{One-shot Teaching} \label{one-shot-teaching}
To realize mobile manipulator process automation in the automation stage, the mobile platform's location information in the pre-built map, the working environment information (colored 3D point cloud of the workbench, tools, etc.) and the robot arm's end-effector path information are collected in real-time through one-shot teaching in this stage \textcolor{black}{in this sub section}.  

In a known workspace (that is, the navigation map has been built beforehand by some SLAM algorithm, such as Gmapping~\citep{gmapping2007}, Cartographer~\citep{cartographer2016} and so on), the mobile manipulator is controlled to finish a task denoted by $tk_{i} (i=1,2,3, ...)$ . Each task $tk_{i}$ contains the mobile platform location information $L_{i}(vehicle) $, the initial working environment information - a colored 3D point cloud $X_{i}(O)$ and the robot end-effector's path $Path_{i}(O)$ when performing the task $tk_{i}$. 
The location information $L_{i}(vehicle) $ is recorded only once, which consists of the position and Euler angle of the mobile platform in the world coordinate system $XYZO_{W}$.
The colored 3D point cloud $X_{i}(O)$ is acquired by the depth sensor mounted on the robot arm in its depth sensor coordinate system $XYZO_{C}$. 
The recorded path information\footnote{We perform one-shot teaching of the desired path by having a human operator manually guiding and applying force on the end-effector on the robot arm. Collaborative robot arms have a zero gravity compensation mode, where the robot exerts just enough torque to compensate for the force applied by gravity. Under this mode, a human operator is able to move and guide a robot by applying force directly on the robot body.} $Path_{i}(O)$ consists of multiple sample points containing information of the pose of the end-effector. The path is recorded with a time frequency of $f_{i}(O)$ Hz. The position and Euler angles of the end-effector are recorded in the robot arm's base coordinate frame $XYZO_{B}$.

After collecting the above data for task $tk_{i}$ only once, the one-shot teaching comes to an end.

\subsection{Mobile manipulator process automation}\label{mobile-manipulator-process-automation}
In the current stage, to finish the task $tk_{i}$, the mobile manipulator process automation needs to accomplish the following four steps: (1)  Firstly, the mobile platform navigates to the location $L_{i}(vehicle)$ and finally parks at the location $\tilde{L}_{i}(vehicle)$ autonomously (\textcolor{black}{Section~\ref{robot-navigation}}). (2) Then, the IPE algorithm calculates the relative 6D parking pose between the location $L_{i}(vehicle)$ and $\tilde{L}_{i}(vehicle)$ in the robot arm's base coordinate frame (\textcolor{black}{Section~\ref{iterative-pose-estimation}}). (3) Next, the adaptive online path learning part learns the difference in pose and modifies the initial robot end-effector path information $Path_{i}(O)$ into the new one $\widetilde{Path}_{i}(arm)$ in the robot arm's base coordinate frame\footnote{$Path_{i}(O)$ and $\widetilde{Path}_{i}(arm)$ are nearly the same in the world coordinate system.} (\textcolor{black}{Section~\ref{subsubsection:adaptive-online-path-learning}}). (4) Finally, the mobile manipulator carries out the task $tk_{i}$ using the newly-learned path $\widetilde{Path}_{i}(arm)$ (\textcolor{black}{Section~\ref{subsubsection:task-excution}}).    

\subsubsection{Robot navigation}\label{robot-navigation}

During the automation stage, the mobile manipulator will attempt to navigate to the desired parking location autonomously $L_{i}(vehicle)$. However due to errors within the system, the robot will eventually park at a different location $\tilde{L}_{i}(vehicle)$ that is close to $L_{i}(vehicle)$. Due to the uneven ground, slippery floor, dynamic obstacles or some other random factors in the real working environment, the mobile platform usually cannot park precisely at the location $L_{i}(vehicle)$. Previous work using the 2D QR code template matching~\citep{iterative-learning-error-compensation2021} could only provide the mobile platform's relative 3D parking pose between the location $L_{i}(vehicle)$ and $\tilde{L}_{i}(vehicle)$. Given the acquired pose from the 2D QR code template matching has only three degrees of freedom, the pose could only be applied to the mobile platform to move on the 2D plane, which couldn't be used to calibrate the robot arm's path in 3D space. To get the extremely-accurate robot arm base's relative 6D pose between the location $L_{i}(vehicle)$ and  $\tilde{L}_{i}(vehicle)$, the iterative 6D pose estimation by eye \& hand interaction will be proposed in the following part.    

\subsubsection{IPE:Iterative pose estimation by eye \& hand}\label{iterative-pose-estimation}
\begin{figure}
	\centering
	\includegraphics[width=0.5\linewidth]{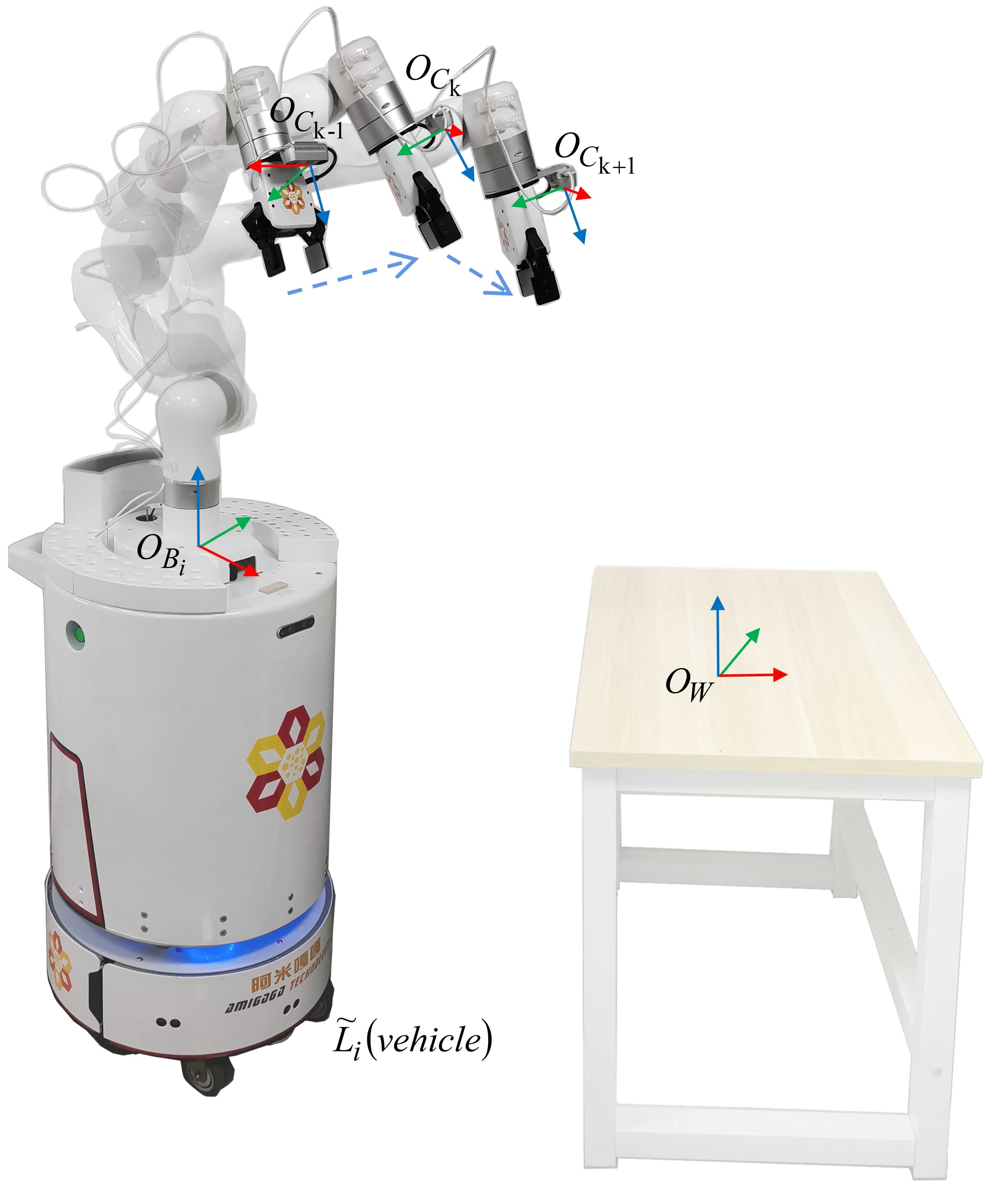}
	\caption{The iterative process by eye-hand interaction to search for the same sampling pose in the teaching stage in world frame.}
	\label{fig:IPE}
\end{figure}

\begin{figure}
	\centering
	\includegraphics[width=0.5\linewidth]{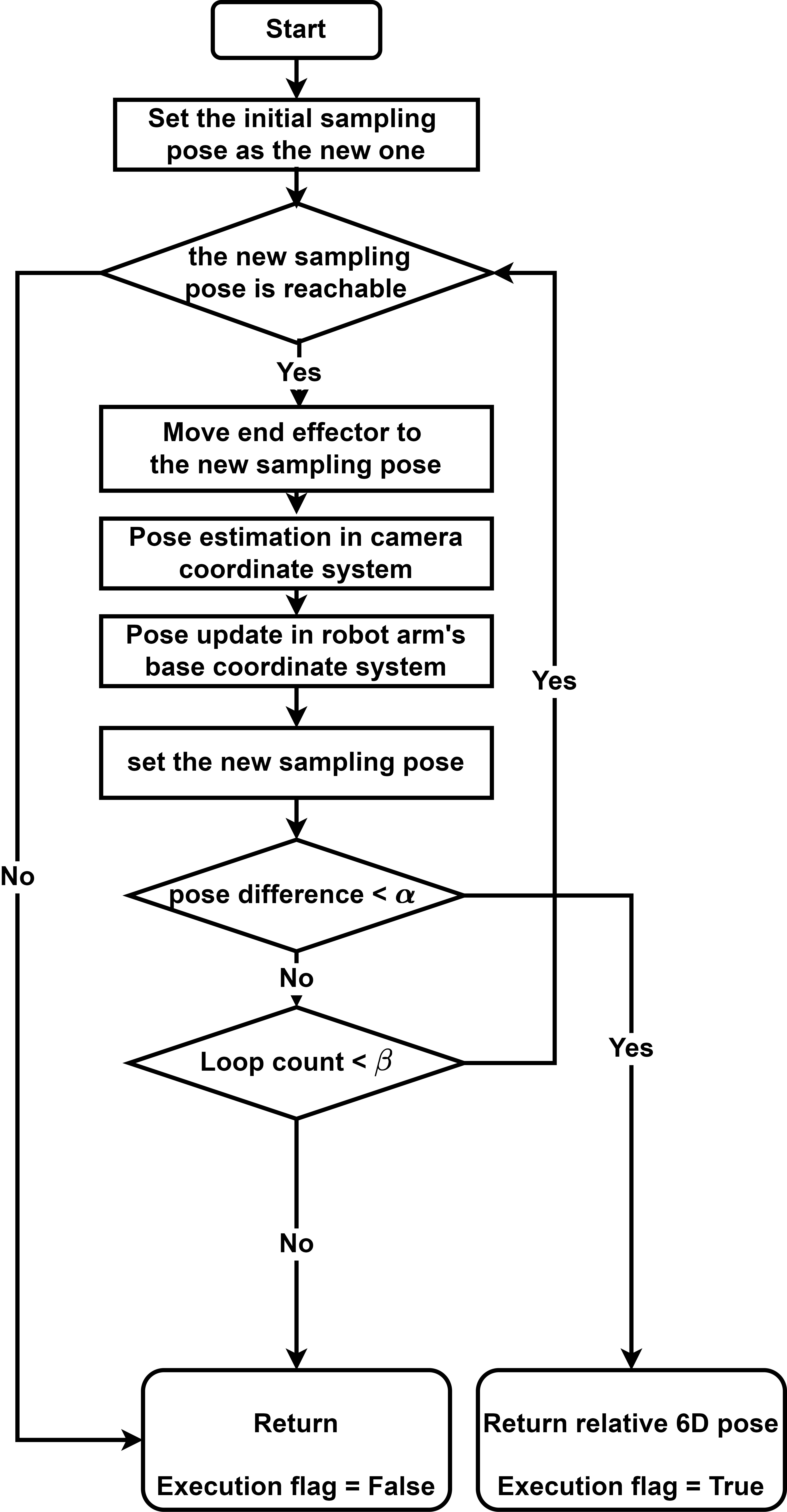}
	\caption{Flowchart of the proposed iterative pose estimation by eye-hand interaction approach (IPE).}
	\label{fig:flowchart}
\end{figure}

Figure~\ref{fig:IPE} shows the iterative process when performing the IPE algorithm. The robot arm's base coordinate system is $XYZO_{B_i}$ when reaching the location $\tilde{L}_{i}(vehicle)$. The robot arm will move the mounted depth camera iteratively to search for the initial sampling pose from the teaching stage. After $k^{th}$ movement of the robot arm, the camera coordinate system becomes $XYZO_{C_k}$.  Figure~\ref{fig:flowchart} illustrates the flowchart of the proposed IPE algorithm.  When the mobile manipulator reaches the working place in the automation mode, the IPE algorithm starts. The robot arm's new sampling pose is set to the initial one in the robot arm base coordinate frame when sampling the working environment's colored 3D point cloud in the one-shot teaching stage. Taking the system safety into account, if the new sampling pose of the end-effector is unreachable by the robot arm, the IPE algorithm will exit with the "False" execution flag. If the new sampling pose of the end-effector is available, the robot arm will move the end-effector to the new sampling pose to sample the working environment's colored point cloud $X_{i}(k), k=1,2,3,...$ ($k$ represents the $k^{th}$ movement of the robot arm in IPE). Then we calculate the relative 6D pose [$\Delta R_{C_{k}}$, $\Delta t_{C_{k}}$] between the sampled point cloud $X_{i}(O)$ from teaching stage and the sampled point cloud $X_{i}(k)$ in the current view in the automation stage through the colored point cloud global registration algorithm (See \textbf{\uppercase\expandafter{\romannumeral2}\quad  Method: colored point cloud global registration} for more details). Then we will update the relative 6D parking pose  [$\Delta R_{B}(k)$, $\Delta t_{B}(k)$] of the robot arm's base (For more details, see \textbf{\uppercase\expandafter{\romannumeral1}\quad \textcolor{black}{Justification}: dynamic pose update in different coordinate systems}). According to the updated relative parking pose [$\Delta R_{B}(k)$, $\Delta t_{B}(k)$], the next new sampling pose is set. If the pose difference [$\Delta R_{C_{k}}$, $\Delta t_{C_{k}}$] is below the threshold $\boldsymbol{\alpha}$, IPE will exit and return the relative 6D pose [$\Delta R_{B}(k)$, $\Delta t_{B}(k)$] in the robot arm's base coordinate system with a ``True" execution flag. If the pose difference [$\Delta R_{C_{k}}$, $\Delta t_{C_{k}}$] is above the threshold $\boldsymbol{\alpha}$ and the loop count is below the threshold $\beta$, IPE will run into the next loop and check if the new sampling pose is reachable or not. If the loop count is above the threshold $\beta$,  the IPE algorithm will exit with the "False" execution flag.        

\noindent \textbf{\uppercase\expandafter{\romannumeral1}\quad \textcolor{black}{Justification}: dynamic pose update in different coordinate systems}

IPE performs iterative pose estimation multiple times, which adjusts the sampling pose of the mounted depth camera incrementally by moving the robot arm to match the initial sampling pose of the camera in the one-shot teaching stage in the world frame. IPE relies on a unique property, that is, if the new sampling pose of the depth camera in the automation stage is closer to the initial sampling pose of the depth camera in one-shot teaching stage in the world frame, the newly sampled point cloud in automation stage will be more similar to the initial sampled point cloud in the one-shot teaching stage. It means there will be more common features for the colored point cloud registration to match thus to get a higher matching accuracy. Thus, in IPE algorithm, the robot arm tends to move the depth camera in an eye-hand iterative style\footnote{Eye-hand iterative style means doing registration and moving the mounted camera by the robot arm iteratively.} for multiple times to position the depth camera in its initial sampling pose in the one-shot teaching stage. \textcolor{black}{Experiments (see Figure~\ref{fig:IPE-performance-against-iteration}) show that with each iteration, the IPE rotation and translation errors gradually decrease. The result from the final iteration is more accurate than the result obtained from the first iteration. }

Due to the eye-hand dynamic iteration process, the robot arm has to change its configuration to reposition the depth camera constantly, therefore the spatial relationship between the depth camera and robot arm's base is constantly changing. In this part, given the dynamic camera pose and robot arm's dynamic movement, we will deduce the robot arm's base relative 6D pose between the teaching stage and the current automation stage. 

\begin{figure}
	\centering
	\includegraphics[width=0.5\linewidth]{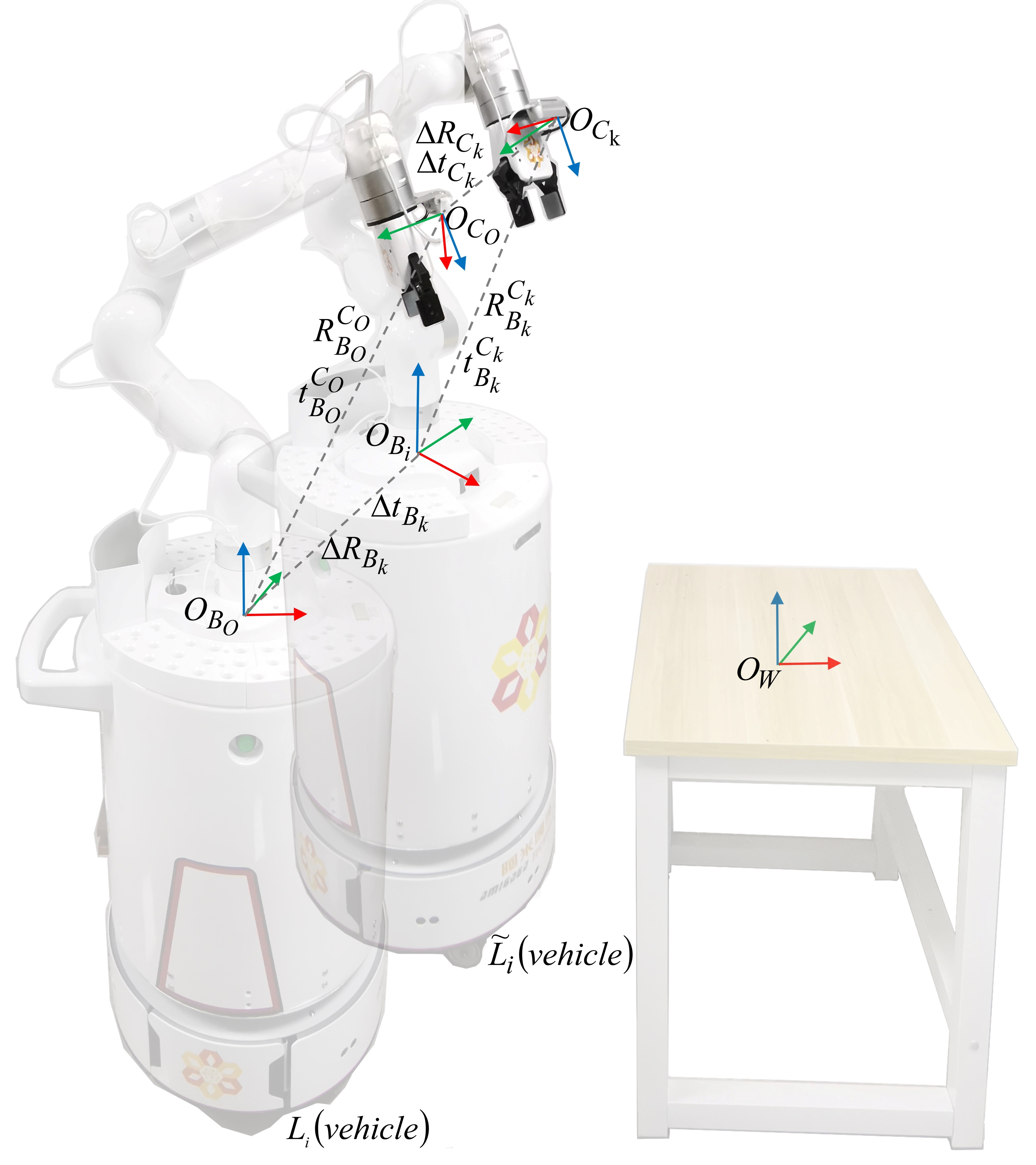}
	\caption{The figure illustrates how the relative parking pose of the robot arm's base can be calculated given the spacial relationship of the camera and base and the relationship between different camera poses.}
	\label{fig:naviation-in-automation}
\end{figure}

Figure~\ref{fig:naviation-in-automation} shows the relative 6D pose between different coordinate systems.
 $R^{C_o}_{B_o}$ and $t^{C_o}_{B_o}$ \textcolor{black}{represent} the rotation and translation matrix from the camera coordinate system to the robot arm's base coordinate system when sampling the working environment's colored point cloud in the one-shot teaching stage in task $tk_{i}$. $R^{C_k}_{B_k}$ and $t^{C_k}_{B_k}$ \textcolor{black}{represent} the rotation and translation matrix from camera coordinate system to robot arm's base coordinate system after the $k^{th}, k=1,2,3, ...$ movement of the robot arm in IPE. $S_{C_o}$ represents one 3D point in camera coordinate system, which corresponds to the 3D point $S_{B_{o}}$ in the robot arm's base coordinate system, in one-shot teaching stage. $S_{C_k}$ represents one 3D point in camera coordinate system, which corresponds to the 3D point $S_{B_{k}}$ in robot arm's base coordinate system, after the $k^{th}$ movement of the robot arm in the automation stage. $\Delta R_{C_k}$ and $\Delta t_{C_k}$ \textcolor{black}{represent} the camera's relative rotation and translation matrix between the teaching stage and the current automation stage after the $k^{th}$ movement of the robot arm. $\Delta R_{B_k}$ and $\Delta t_{B_k}$ \textcolor{black}{represent} the robot arm base's estimated relative rotation and translation matrix when sampling working environment's point cloud in one-shot teaching stage and after the $k^{th}$ movement of the robot arm during the iterative adjustment process.  

\begin{align}
R^{C_o}_{B_o}S_{C_o}+t^{C_o}_{B_o}&=S_{B_o} \label{eq:11} \\
R^{C_k}_{B_k}S_{C_k}+t^{C_k}_{B_k}&=S_{B_k} \label{eq:12} \\
\Delta R_{C_k}S_{C_o}+\Delta t_{C_k} &= S_{Ck} \label{eq:13}\\
\Delta R_{B_k}S_{B_o}+\Delta t_{B_k} &= S_{B_k} \label{eq:14} 
\end{align}

From equation \eqref{eq:11} and \eqref{eq:12} we get
\begin{align}
S_{C_o} &= R^{C_o-1}_{B_o}S_{B_o}- R^{C_o-1}_{B_o}t^{C_o}_{B_o} \label{eq:15}\\
S_{C_k} &= R^{C_k-1}_{B_k}S_{B_k}-R^{C_k-1}_{B_k}t^{C_k}_{B_k} \label{eq:16}
\end{align}

Substitute $S_{C_o}$ and $S_{C_k}$ in equation \eqref{eq:13} with equation \eqref{eq:15} and \eqref{eq:16}. We get

\begin{equation}
\begin{split}
&\Delta R_{C_k}\left( R^{C_o-1}_{B_o}S_{B_o}-R^{C_o-1}_{B_o}t^{C_o}_{B_o}\right)+\Delta t_{C_k} =\\
&R^{C_k-1}_{B_k}S_{B_k}-R^{C_k-1}_{B_k}t^{C_k}_{B_k} 
\end{split}   \label{eq:17}
\end{equation} 

Reformulating the equation~\eqref{eq:17} by multiplying $R^{C_k}_{B_k}$ on the both sides, we get:
\begin{equation}
\begin{split}
&R^{C_k}_{B_k}\Delta R_{C_k}R^{C_o-1}_{B_o}S_{B_o}+ (-R^{C_k}_{B_k}\Delta R_{C_k}R^{C_o-1}_{B_o}t^{C_o}_{B_o}\\
&+R^{C_k}_{B_k}\Delta t_{C_k}+t^{C_k}_{B_k}) = S_{B_k}
\end{split}  \label{eq:18}
\end{equation}	

Comparing equation~\eqref{eq:18} and equation~\eqref{eq:14}, we get:
\begin{equation}
\Delta R_{B_k} = 	R^{C_k}_{B_k}\Delta R_{C_k}R^{C_o-1}_{B_o} \label{eq:19}
\end{equation}	
\begin{equation}
\Delta t_{B_k} = -R^{C_k}_{B_k}\Delta R_{C_k}R^{C_o-1}_{B_o}t^{C_o}_{B_o}+R^{C_k}_{B_k}\Delta t_{C_k}+t^{C_k}_{B_k}
\label{eq:20}
\end{equation}	

\noindent \textbf{\uppercase\expandafter{\romannumeral2}\quad  Method: colored point cloud global registration}

In order to get the mounted camera's relative rotation matrix $\Delta R_{C_k}$ and translation matrix $\Delta t_{C_k}$, the colored point cloud global registration method\footnote{Readers could also use other 3D point cloud registration algorithms to replace the colored point cloud global registration in this part as well.} is proposed in this part by simply combining the fast global registration~\citep{fast-global-registration2016} and the local colored point cloud registration~\citep{colored-point-cloud-registration-2017} with some improved engineering techniques (e.g.: point cloud pre-processing, parameter fine-tuning, multi-scale matching, etc.). More specifically, the coarse global pose from  the fast global registration~\citep{fast-global-registration2016} is fed into the colored point cloud registration~\citep{colored-point-cloud-registration-2017} algorithm to avoid the local minimum during the registration. The local algorithm~\citep{colored-point-cloud-registration-2017} takes not only the geometry information but also the color information into the account to achieve a better registration accuracy. The proposed method with the simple strategy above provides a robust and accurate 6D pose estimation for IPE framework, which meets the mobile manipulator process automation's localization requirement.

\subsubsection{Adaptive online path learning} \label{subsubsection:adaptive-online-path-learning}

The robot end-effector's path for performing task $tk_{i}$ is recorded in the robot arm's base frame during one-shot teaching. When the robot reaches the same working place again in the automation stage, the robot arm base's pose will differ from that of the one-shot teaching stage because of motion or system errors or some other random factors. To ensure the robot end-effector follows the desired path in the world frame to finish the manipulation task $tk_{i}$, the robot has to learn to adjust the reference path ${Path}_{i}(O)$ obtained in teaching stage to the new path $\widetilde{Path}_{i}(arm)$ in the robot arm's base coordinate frame in the automation stage.

Considering a sampled point $S_{B_o}$ from the recorded path ${Path}_{i}(O)$ in the teaching stage, the position of the end-effector is transferred from the teaching stage  to the automation stage in the robot arm's base coordinate system using the following equation
\begin{align}
\Delta R_{B_k}S_{B_o}+\Delta t_{B_k} &= S_{B_k} \label{eq:21}
\end{align}
where $\Delta R_{B_k}$ and $\Delta t_{B_k}$ could be calculated using equation~\eqref{eq:19} and equation~\eqref{eq:20}. The sampled point $S_{B_k}$ is from the new execution path $\widetilde{Path}_{i}(arm)$ in automation stage.

\subsubsection{Task Execution} \label{subsubsection:task-excution}
After obtaining the newly-adjusted robot end-effector path  $\widetilde{Path}_{i}(arm)$, some motion planning algorithm (e.g.: RRT*~\citep{RRT2011}, PRM~\citep{PRM1996}) could be used to compute the trajectory containing the target joint angles which are used to control the robot joints during execution. The acceleration, torque, velocity, and position limit of the robot joints are taken into account by the motion planning algorithm. \textcolor{black}{If the motion planing algorithm could compute a valid trajectory, the robot will execute the task. If not, the robot will abandon this task.}

\section{Experiment}\label{Experiment}

\subsection{Experiment Configuration}\label{experiment-configuration}

\begin{figure*}
	\begin{center}
		\subfloat{\includegraphics[width=1\linewidth]{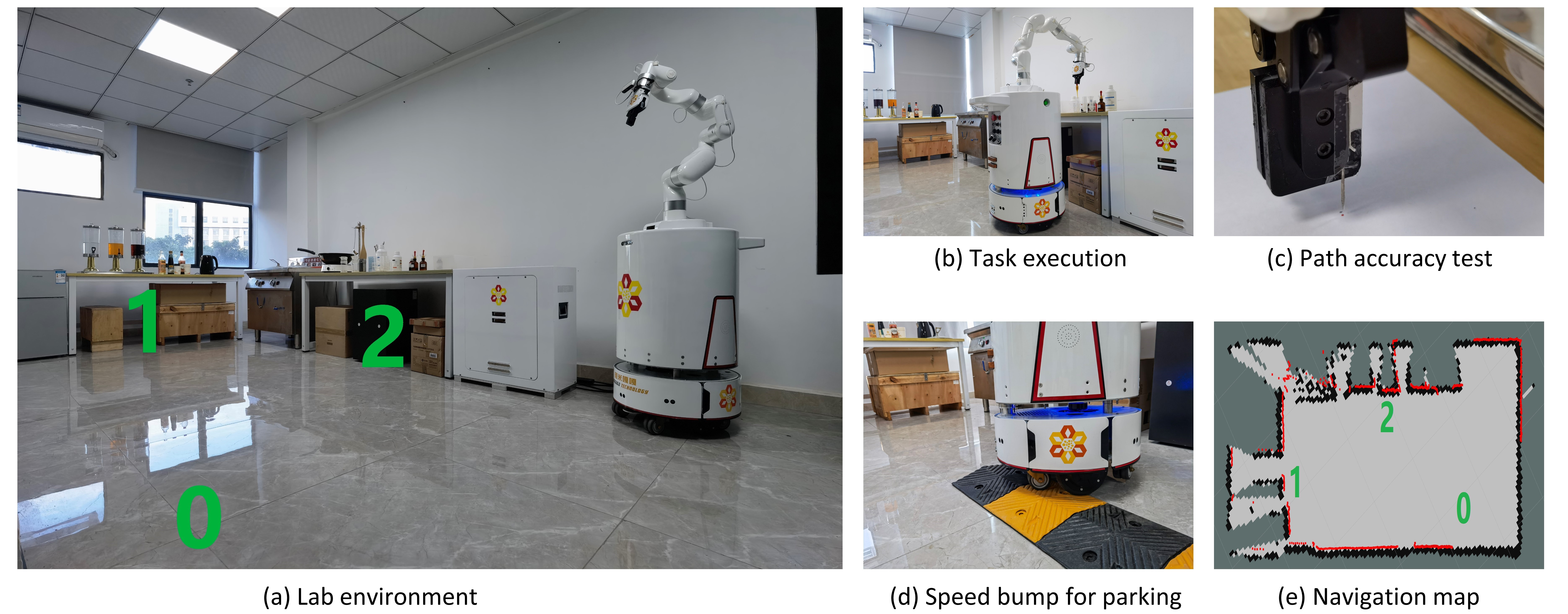}}	
	\end{center}
	\caption{Experiment configuration. (a) shows the experiment environment. The mobile manipulator will conduct experiments at site 0, site 1, site 2, site 2' (site 2 and site2' are in the same location). (b) shows the robot executing a task.  (c) shows the executed path's accuracy test using a needle. (d) shows a speed bump installed at the parking place to emulate uneven terrain. (e) shows the pre-built map for robot navigation. \textcolor{black}{The black lines or dots in the occupancy map represent obstacles ("occupied"). The red lines or dots represent the laser scan which is used for localization. '0', '1' and '2' represent site 0, site 1 and site 2.}}
	\label{fig:experiment-scene}
\end{figure*}

The mobile manipulator "Gagabot DR-03"\footnote{For more details, please refer to \url{http://www.amigaga.com/en/index.php?id=111}} is the platform to demonstrate the performance of the proposed method.  Gagabot DR-03 consists of a two-wheeled differential mobile platform and a 6-DOF (\textcolor{black}{Degree of Freedom}) robot arm. A depth camera \textcolor{black}{"Intel Realsense D435"} is mounted at the end of the robot arm to perceive the environment. Inside Gagabot DR-03, an integrated industrial computer (Nvidia AGX Xavier) is used for the complex computation (e.g.: navigation, perception, planning, control, etc.). The Nvidia AGX Xavier (105 mm x 105 mm x 65 mm) is equipped with a 512-core Volta GPU, a 8-core ARM v8.2 64-bit CPU and 32GB memory. All the experiments in this paper were conducted with the Nvidia AGX Xavier (See Figure~\ref{fig:experiment-scene}).  Figure~\ref{fig:experiment-scene}, (a) shows an image of the lab where we conduct the experiments. The robot will navigate from Site 0 to Site 1, Site 2, Site 2' to perform the experiments (Site 2 and Site2' are in the same location). (b) shows an example of the robot executing a task.  (c) shows the executed path's accuracy test by using a needle. (d) shows a speed bump installed on the parking position to emulate uneven terrain. (e) shows the pre-built map from Gmapping~\citep{gmapping2007} for the robot navigation. 

The experiments are divided into 3 groups given different working sites (Site 1, Site 2, Site 2'). Three different kinds of experiments and analysis are conducted for each group: 1) parking pose accuracy in Section~\ref{parking-pose-accuracy}, 2) IPE performance in Section~\ref{IPE-performance} and 3) executed path's accuracy in Section~\ref{excuted-path-accuracy}. Site 1\footnote{\textcolor{black}{Site 1 consists of cups, big liquid containers with valves, small bottles containing different liquid, desks, logo and letters printed on the tape.}} emulates a plant extract workshop to make the robot perform complex manipulation with a flat floor, such as switching the valve, fetching and returning the cup. Site 2\footnote{\textcolor{black}{Site 2 consists of glass and plastic reagent bottles (containing sodium hydroxide, hydrochloric acid, phenolphthalein, ferrum, seperately) , droppers, a waste disposal container, test tubes, a test tube stand, a desk, logo and letters printed on the tape.}} emulates a quality inspection center in a factory with a flat floor to test the liquid products by chemical reaction. Site 1 and Site 2 consists of only flat floor. However, in the real world, there are many harsh working environments where the floor is uneven. Taking real-world floor conditions into consideration, a speed bump\footnote{\textcolor{black}{The speed bump's shape is a triangular prism, with the size $1000mm * 350mm * 50mm $ ($length * width * height$). It weighs 11 $kg$ and has a surface with uneven 3D texture for increased friction. }} is installed at site 2 to emulate uneven terrain. The new site is denoted as site 2'. 
Thus, we could get various 6D parking poses on a 3D surface for the experiments. We define this group of experiments with a speed bump to park on (see Figure~\ref{fig:experiment-scene} d) are at Site 2'. Each experiment is conducted 30 times. For an explicit overview of all the experiments, please see Table~\ref{table:count-experiments}. We have used two formats to describe the 6D pose. The first way is using a 6D vector $[tx, ty, tz, r, p, y]$ (short for [translation on $x$ axis, translation on $y$ axis, translation on $z$ axis, $roll$, $pitch$, $yaw$]). The unit is meter for $tx$, $ty$, $tz$ and degree for $r$, $p$, $y$. We use the difference between the estimated 6D vector and the ground truth to describe the 6D pose's error. The second way is using rotation matrix $\boldsymbol{R}$ and translation matrix $\boldsymbol{t}$. The equations for estimating the accuracy of the 6D poses are from~\citep{Metrics_3D_Rotation_huynh2009}:

\begin{align}
E_R = ||\boldsymbol{I-R_{gt}R_{est}^{-1}}||_{F} \label{eq:23}
\end{align}
\begin{align}
E_t = ||\boldsymbol{t_{gt}-t_{est}}||_{F}
\label{eq:24}
\end{align}
where $\boldsymbol{t_{est}, R_{est}}$ are the estimated values and $\boldsymbol{t_{gt}, R_{gt}}$  are the ground truth respectively.  $|| \bullet || _{F}$ is the Frobenius norm. Readers should know that: Although the error estimation methods and their error value for the two formats above are different, they all could describe the 6D pose's accuracy.  

\begin{table}[h!]
	\centering
	\caption{Number of Different Experiment Trials} \label{table:count-experiments}
	\begin{tabular}{|p{1.5cm}|| p{1.5cm}| p{1.5cm}| p{1.5cm}|}
		\hline 
		\textit{Factor}  &  \textit{Site 1} & \textit{Site 2}  & \textit{Site 2'}\\
		\hline 
		parking pose accuracy  &   30 times & 30 times & 30 times \\
		\hline 
		IPE performance & 30 times & 30 times & 30 times \\ 
		\hline 
		executed path's accuracy & 30 times & 30 times & 30 times\\ 
		\hline 
	\end{tabular} 	
\end{table}

\subsection{Parking Pose Accuracy}\label{parking-pose-accuracy}
Analysis is performed to observe if there is any relationship between the parking pose's accuracy and IPE performance. In each experiment, the robot will be required to navigate to a specified location. To get the ground truth of the relative parking pose, the end-effector is pointed to a fixed point with a fixed pose when the mobile base parks at the specified location. The transformation matrix from the robot arm's base to the fixed point is $T_0$. When the mobile base navigates to the specified location again, we let the end-effector point to that fixed same point with that same pose. The current transformation matrix from the robot arm's base to the fixed point is $T_1$. Given the robot arm is extremely accurate with a repeatability of 0.1mm, we use the product of those transformation matrixes $T_0 T^{-1}_{1}$ to represent the ground truth of the relative parking pose. 

We designate the position (2.0415, -1.375, 0.0) with orientation (0.0, 0.0, -0.2970, 0.9548) (represented by \textcolor{black}{unit} quaternion) as Site 1 in the map ( See Figure~\ref{fig:experiment-scene} e ).  We specify the position (0.1251, -0.7638, 0.0) with orientation (0.0, 0.0, -0.8785, 0.4776) (represented by \textcolor{black}{unit} quaternion) as Site 2. The starting point's position (site 0) is (-0.3498, 0.9449, 0.0) with a orientation (0.0, 0.0, -0.5367, 0.8437). The mean and standard deviation of the relative parking pose are listed in Table~\ref{table:parking-pose-accuracy} by using the 6D vector $[tx, ty, tz, r, p, y]$ format. 
If we use Equation~\eqref{eq:23} and Equation~\eqref{eq:24} to describe the error, 
the mean rotation error is 0.0617 at Site 1, 0.0614 at Site 2, 0.2910 at Site 2'. The standard deviation of the rotation error is 0.0418 at Site 1, 0.0433 at Site 2, 0.2213 at Site 2'. The mean translation error is 0.1018 m at Site 1, 0.0799 m at Site 2, 0.1304 m at Site 2'. The standard deviation of the translation error is 0.0176 m at Site 1, 0.0316 m at Site 2, 0.0594 m at Site 2'. 
Note that this is the uncorrected initial base positioning error.

\begin{table}[h!]
	\centering
	\caption{Initial Parking Pose Accuracy} \label{table:parking-pose-accuracy}
	\begin{tabular}{|p{1.5cm}|| p{1.5cm}| p{1.5cm}| p{1.5cm}|}
		\hline 
		\textit{Factor}  &  \textit{Site 1} & \textit{Site 2}  & \textit{Site 2'}\\
		\hline 
		$\overline{tx}$ / (m)  &   0.0011 & 0.0505 & 0.0779 \\
		\hline 
		$\sigma_{tx}$ / (m) & 0.0281 & 0.0287 & 0.0569 \\ 
		\hline 
		$\overline{ty}$ / (m) & -0.0995 & 0.0539 & 0.0261\\ 
		\hline 
		$\sigma_{ty}$ / (m) &   0.0179 & 0.0287 & 0.1060 \\
		\hline 
		$\overline{tz}$ / (m) & -0.0012 & -0.0011 & -0.0042 \\ 
		\hline 
		$\sigma_{tz}$ / (m) & 0.0013 & 0.0014 & 0.0054\\ 
		\hline 
		$\overline{r}$ / (deg) &   0.2151 & 0.1155 & 0.0079 \\
		\hline 
		$\sigma_{r}$ / (deg) & 0.1593 & 0.0784 & 1.2595 \\ 
		\hline 
		$\overline{p}$ / (deg) & -0.0036 & -0.1733 & 0.7932\\ 
		\hline 
		$\sigma_{p}$ / (deg)  &   0.1439 & 0.1295 & 2.6967 \\
		\hline 
		$\overline{y}$ / (deg) & 0.0023 & 1.6706 & -1.9701 \\ 
		\hline 
		$\sigma_{y}$ / (deg) & 2.9214 & 2.7079 & 14.7211\\ 	
		\hline	
	\end{tabular} 	
\end{table}

\subsection{IPE Performance}\label{IPE-performance}

From Equation~\ref{eq:19} and Equation~\ref{eq:20}, the relative parking pose ($\Delta R_{B_k}$, $\Delta t_{B_k}$) and the relative camera pose ($\Delta R_{C_k}$, $\Delta t_{C_k}$) could be deduced or calculated from each other. Thus, we will only test the accuracy of the relative parking pose ($\Delta R_{B_k}$, $\Delta t_{B_k}$) in this part. The method for obtaining the ground truth of relative parking pose is the same in Section~\ref{parking-pose-accuracy}. A threshold has to be set to indicate convergence. We set the convergence threshold $\boldsymbol{\alpha}$ of the sampling pose difference $(x, y, z, roll, pitch, yaw)$ as (0.002m, 0.002m, 0.002m, $0.5^\circ$, $0.5^\circ$, $0.5^\circ$). The maximum iteration number $\beta$ is set as 5 for Site 1, Site 2 and 10 for Site 2' because the situation at Site 2' is harder than those at Site 1, Site 2. This is because the uneven terrain emulated by the speed bump will introduce more changes for point cloud matching in the $z$ axis, roll and pitch orientation, compared to the flat terrain case of Site 1 and Site 2.  

\begin{figure}
	\begin{center}
		\subfloat{\includegraphics[width=1\linewidth]{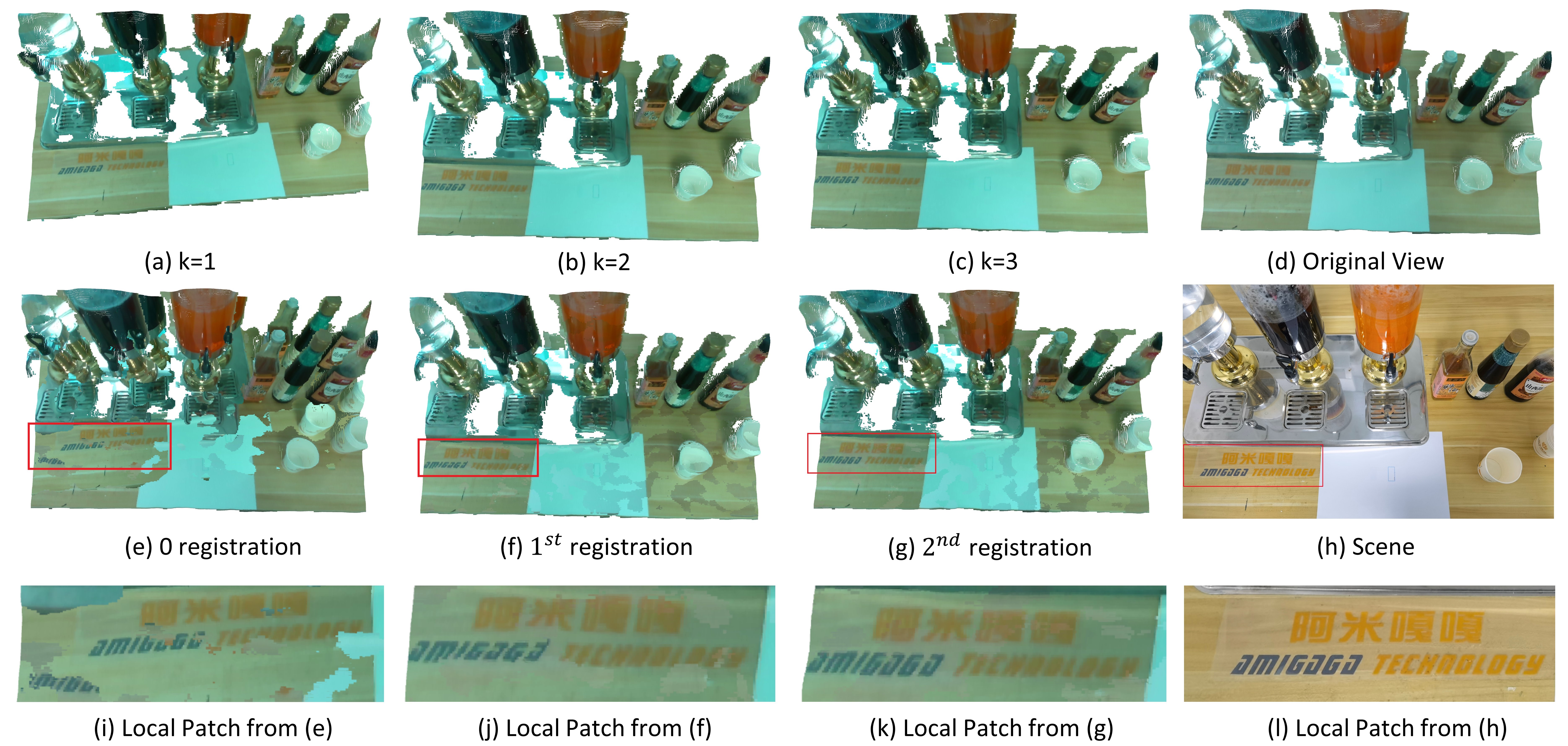}}	
	\end{center}
	\caption{IPE process. (a) - (c) show the point clouds (from the depth camera) after $k^{th}$ movement of the robot arm in IPE. (d) is the point cloud sampled during teaching stage. (e) shows the scene before matching the point cloud (a) and (d). (f) shows the scene before matching point cloud (b) and (d). (g) shows the scene before matching the point cloud (c) and (d). 
	(h) is the RGB image of the scene. (i) - (l) are the local \textcolor{black}{pathes} from (e) - (h).\textcolor{black}{(Readers are encouraged to view the electronic version of the paper for clearer visual details.)}}
	\label{fig:IPE-process-point-cloud}
\end{figure}

Figure~\ref{fig:IPE-process-point-cloud} shows one example of the IPE matching. In Figure~\ref{fig:IPE-process-point-cloud}, (a) - (c) show the point clouds (from the depth camera) after $k^{th}$ movement of the robot arm in IPE. (d) is the point cloud sampled during teaching stage. (e) shows the scene before matching the point cloud (a) and (d). (f) shows the scene before matching point cloud (b) and (d). (g) shows the scene before matching the point cloud (c) and (d). Note the improved alignment. (h) is the RGB image of the scene. (i) - (l) are the local \textcolor{black}{patches} from (e) - (h). With each iteration $k$, the scene [(a), (b) (c)] gets closer to the original (d). That is, the current depth camera pose gets closer iteratively to the sampling pose from the teaching stage. With each additional iteration of the IPE, the registration result [(e),(f),(g)] gradually improves. (i) - (k) shows the details of the registration's accuracy. Compare the word "AMIGAGA TECHNOLOGY" in (j) and (k), some parts are missing in (j). 

In Table~\ref{table:IPE-accuracy}, the mean 
$[\overline{\Delta{tx}}, \overline{\Delta{ty}}, \overline{\Delta{tz}}, \overline{\Delta{r}}, \overline{\Delta{p}}, \overline{\Delta{y}}]$ and standard deviation 
$[\sigma_{{\Delta{tx}}}, \sigma_{{\Delta{ty}}}, \sigma_{{\Delta{tz}}}, \sigma_{{\Delta{r}}}, \sigma_{{\Delta{p}}}, \sigma_{{\Delta{y}}}]$  of the error between the IPE estimation ($\Delta R_{B_k}$, $\Delta t_{B_k}$) and ground truth are given. While using the rotation matrix and translation matrix criteria shown in Equation \eqref{eq:23} and Equation \eqref{eq:24} as error measures, the mean rotation error between IPE estimation and ground truth is 0.0169 at Site 1, 0.0183 at Site 2, 0.0225 at Site 2'. The standard deviation of the rotation error is 0.0074 at Site 1, 0.0063 at Site 2, 0.0103 at Site 2'. The mean translation error is 0.0049 m at Site 1, 0.0054 m at Site 2, 0.0089 m at Site 2'. The standard deviation of the translation error is 0.0025 m at Site 1, 0.0019 m at Site 2, 0.0044 m at Site 2'. From the results above, it can be seen that IPE exhibits high accuracy.

\begin{table}[h!]
	\centering
	\caption{The error of IPE's estimation ($\Delta R_{B_k}$, $\Delta t_{B_k}$) } \label{table:IPE-accuracy}
	\begin{tabular}{|p{1.5cm}|| p{1.5cm}| p{1.5cm}| p{1.5cm}|}
		\hline 
		\textit{Factor}  &  \textit{Site 1} & \textit{Site 2}  & \textit{Site 2'}\\
		\hline 
		$\overline{\Delta{tx}}$ / (m) &   0.0001 & 0.0000 & -0.0004 \\
		\hline 
		$\sigma_{{\Delta{tx}}}$ / (m) & 0.0004  & 0.0002 & 0.0005 \\ 
		\hline 
		$\overline{\Delta{ty}}$ / (m) & 0.0009 & 0.0010 & -0.0005\\ 
		\hline 
		$\sigma_{{\Delta{ty}}}$ / (m) &   0.0010 & 0.0009 & 0.0007 \\
		\hline 
		$\overline{\Delta{tz}}$ / (m) & 0.0007 & -0.0009 & -0.0010 \\ 
		\hline 
		$\sigma_{{\Delta{tz}}}$ / (m) & 0.0012 & 0.0008 & 0.0010 \\ 
		\hline 
		$\overline{\Delta{r}}$  / (deg) &   0.0190 & -0.0106 & 0.0088 \\
		\hline 
		$\sigma_{{\Delta{r}}}$ / (deg) & 0.1472 & 0.0123 & 0.0469 \\ 
		\hline 
		$\overline{\Delta{p}}$ / (deg) & 0.0495 & -0.1218 & -0.1046\\ 
		\hline 
		$\sigma_{{\Delta{p}}}$ / (deg) &   0.1317 & 0.1119 & 0.1090 \\
		\hline 
		$\overline{\Delta{y}}$ / (deg) & -0.1338 & -0.1423 & 0.0311 \\ 
		\hline 
		$\sigma_{{\Delta{y}}}$ / (deg) & 0.1413 & 0.1221 & 0.0744 \\
		\hline		
	\end{tabular} 	
\end{table}

\begin{figure*}
	\begin{center}
	\subfloat{\includegraphics[width=1\linewidth]{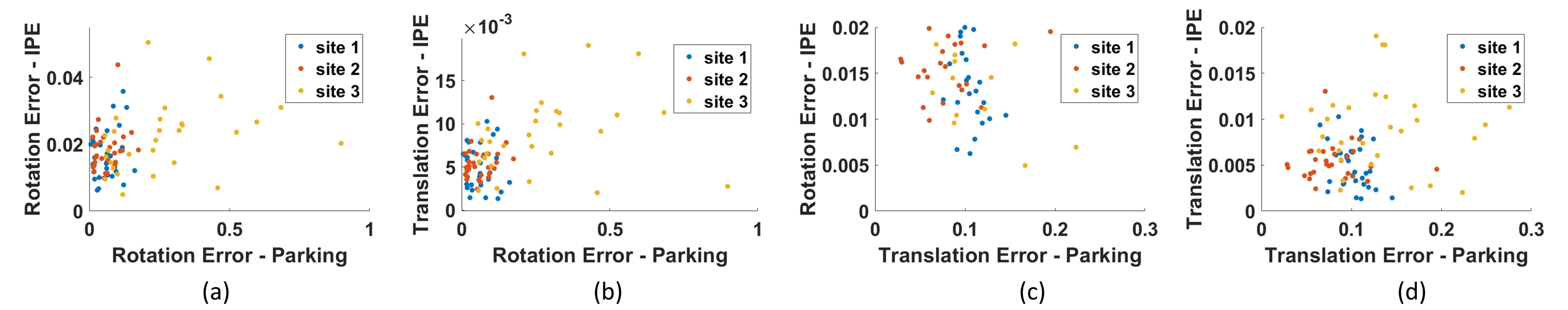}}	
	\end{center}
	\caption{Error Relationship between Parking and IPE.}
	\label{fig:error-relationship-between-parking-and-IPE}
\end{figure*}
	
Figure~\ref{fig:error-relationship-between-parking-and-IPE} shows the error relationship between the actual parking pose and IPE's estimation. Figure~\ref{fig:error-relationship-between-parking-and-IPE} (a) shows the error relationship between the actual parking's rotation and the estimated rotation from IPE. Figure~\ref{fig:error-relationship-between-parking-and-IPE} (b) shows the error relationship between the actual parking's rotation and the estimated translation (unit: m) from IPE.
Figure~\ref{fig:error-relationship-between-parking-and-IPE} (c) shows the error relationship between the actual parking's translation (unit: m) and the estimated rotation from IPE.
Figure~\ref{fig:error-relationship-between-parking-and-IPE} (d) shows the error relationship between the actual parking's translation (unit: m) and the estimated translation (unit: m) from IPE. 
	
Generally, when doing point cloud registration,  with every increment of the initial rotation and translation misalignment, the 6D pose estimation error will increase as well. However, with IPE, this is not the case. From Figure~\ref{fig:error-relationship-between-parking-and-IPE}, we find that the initial rotation and translation change resulting from the parking pose doesn't influence the IPE estimation (i.e. no linear relationship). No matter what the initial rotation and translation is, the IPE's translation estimation error is always below 0.02m and its rotation estimation error is always below 0.02. Note that the IPE error is considerably smaller than the `Parking' error in the Section~\ref{parking-pose-accuracy}. The reason is that IPE has used an iterative way to move the robot arm to sample multiple point clouds at different views until the sampling pose is nearly the same with that at the teaching stage. 
The final accuracy of IPE will be influenced by the 6D pose estimation method inside IPE (e.g.: we use the colored point cloud registration in this paper). Given that the colored point cloud global registration is proposed in other work~\citep{colored-point-cloud-registration-2017,fast-global-registration2016} and is not the contribution of this paper, experiments for testing the performance of the global colored point cloud registration are omitted. Readers could be directed to the original two papers~\citep{colored-point-cloud-registration-2017,fast-global-registration2016} for more information. 
From this part, the experiment shows the IPE's strong robustness and accuracy, which is not influenced by the initial rotation and translation.

\begin{figure*}
	\begin{center}
		\subfloat{\includegraphics[width=1\linewidth]{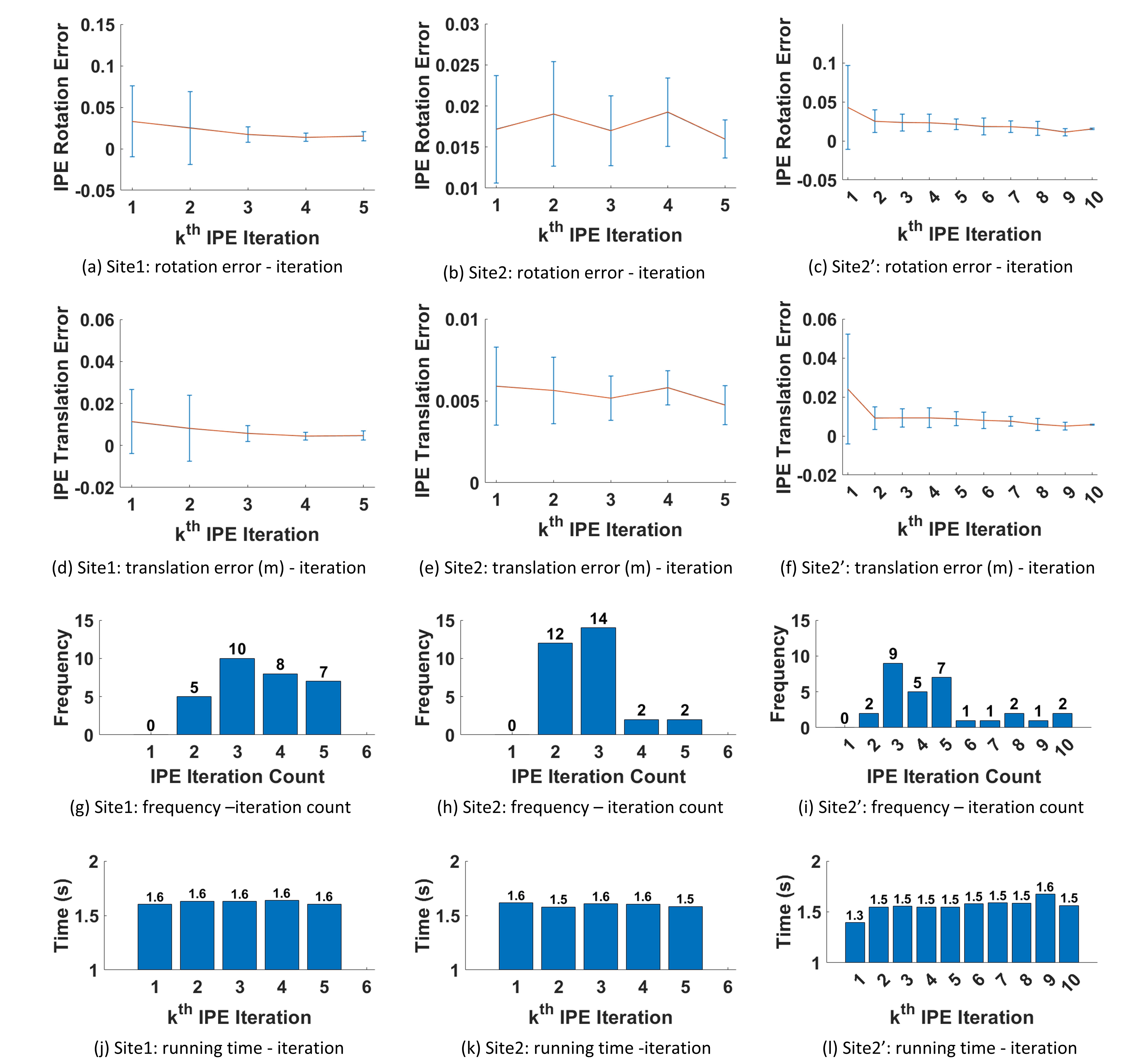}}	
	\end{center}
	\caption{IPE performance within each iteration.}
	\label{fig:IPE-performance-against-iteration}
\end{figure*}
	
Figure~\ref{fig:IPE-performance-against-iteration} shows the performance of the IPE algorithm at each iteration. Figure~\ref{fig:IPE-performance-against-iteration} (a) - (c) shows the IPE's estimated rotation error at each iteration for Site 1, Site 2, Site 2'. (d) - (f) shows the IPE's estimated translation error at each iteration for Site 1, Site 2, Site 2'. (g) - (i) shows the distribution of the number of iterations to convergence at Site 1, Site 2, Site 2'. (j) - (l) shows the average running time for registration at $k^{th}$ iteration. From (a) - (f), we find that with every increasing iteration, both the mean and standard deviation of the rotation error and translation error gradually decreases, indicating an increase in accuracy and robustness. 
\textcolor{black}{
	The reason is that IPE adjusts the sampling pose of the mounted depth camera incrementally by moving the robot arm to match the initial sampling pose of the camera in the one-shot teaching stage in the world frame. Thus, the newly sampled point cloud in the latest iteration in the automation stage will be more similar to the initial sampled point cloud in the one-shot teaching stage. It means there will be more common features for the colored point cloud registration to match to get a higher matching accuracy.
} 
From (g) - (i), we find that IPE needs at least 2 iterations to converge and the majority of the experiments converge by  iteration 3-5.
\textcolor{black}{The result from the final iteration (in which IPE converges) is more accurate than the result obtained from the first iteration.
It proves that the eye-hand iterative strategy in IPE could efficiently reduce the pose estimation error by incrementally positioning the depth camera closer to its initial sampling pose in the one-shot teaching stage.	
}
 From (j) to (l), we find that the average running time of each registration is stable, usually ranging from 1.5s - 1.7s. 
 \textcolor{black}{It shows IPE's strong robustness against the rotation and translation perturbation.}
  In future work, GPU-based acceleration could be implemented to \textcolor{black}{speed up the point cloud registration algorithm inside IPE in each iteration}.

All the experiments on Site 1, Site 2, Site2' are successful and don't exit from IPE algorithm with failure (see Figure~\ref{fig:flowchart}, "Execution flag = False" ) because the robot arm's operating range (radius = 0.8 m) is big enough to compensate for the parking error. If the robot parks too far away from the taught position (e.g. > 0.8 meter) and the robot arm fails to find a feasible solution to move the depth camera to the desired pose in the next iteration, IPE will abort with failure definitely. If the colored point cloud global registration could not get converged to make the pose difference fall below the threshold $\boldsymbol{\alpha}$ even when the IPE's loop count exceeds the maximum iteration number $\beta$, IPE will abort with failure as well. Given that the colored point cloud global registration is proposed in other work~\citep{colored-point-cloud-registration-2017,fast-global-registration2016} and is not the contribution of this paper, experiments for testing the performance of the global colored point cloud registration are omitted. \textcolor{black}{The two original papers~\citep{colored-point-cloud-registration-2017,fast-global-registration2016} have given their performance (robustness and accuracy) against noise, density, occlusion, overlapping rate, rotation and translation perturbation of the input point clouds.} 

\subsection{Executed Path's Accuracy}\label{excuted-path-accuracy}
\begin{figure*}
	\begin{center}
		\includegraphics[width=\linewidth]{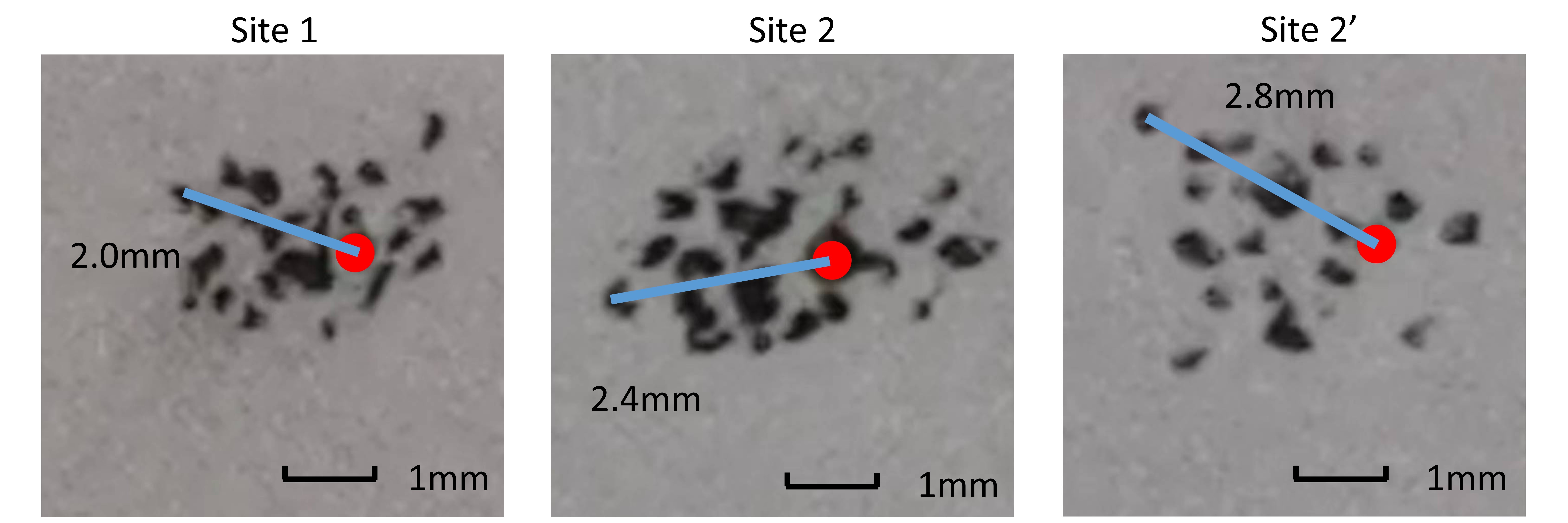}
	\end{center}
	\caption{The record of needle tip's positions at different sites. (10x magnification) \textcolor{black}{(Readers are encouraged to view the electronic version of the paper for clearer visual details.)}}
	\label{fig:needle-tip}
\end{figure*}

In this part, the position accuracy of the end-effector's path $\widetilde{Path}_{i}(arm)$  will be tested. A needle is attached to the end-effector (See Figure~\ref{fig:experiment-scene} c). The needle tip is pointed at a specified point on the paper (recorded by a red dot) during the teaching stage and the needle tip is required to point to the same point during the automation stage. The bias of the needle tip's position (recorded by a black dot at each trial) from the specified target point (recorded by a red dot) over 30 trials during automation stage is used to show the accuracy of the executed path's position.  Figure~\ref{fig:needle-tip} records the needle tip's positions in different trials. 
Because the recorded dots crowd in a very small area (e.g.: a small circle with 2 - 3 mm radius) it is not easy to calculate the distance between each black dot and the red dot accurately enough. Thus, we only measure the maximum distance bias, the mean distance bias and its standard deviation to describe the position accuracy of the point on the executed path in a rough and approximated manner. According to the experiment results (maximum distance bias is 2.0 mm at Site 1, 2.4 mm at Site 2, 2.8 mm at at Site 2'; mean distance bias is 1.0 mm at Site 1, 1.1 mm at Site 2, 1.2 mm at at Site 2'; standard deviation of the distance bias is 0.5 mm at Site 1, 0.5 mm at Site 2, 0.6 mm at at Site 2'), the end-effector's position accuracy could meet the accuracy requirement of the most applications in daily life.
\textcolor{black}{The factors that contributes to the executed path's high accuracy are the high pose estimation accuracy of IPE and the high repeatability of the collaborative robot arm ($\pm 0.1$mm).} 
This experiment only measures the position accuracy of the executed path, because the orientation of the end-effector\footnote{Except the position of the end-effector, the rest parameters of the robot arm's joints are not fixed and calculated by some motion planning algorithm (e.g.: RRT*~\citep{RRT2011}, PRM~\citep{PRM1996}).} is not as important as the position for many manipulation tasks.

\section{\textcolor{black}{Discussion}} \label{Discussion}

\textcolor{black}{
	In this work, we collect human demonstration data for one-shot teaching through kinesthetic guidance, i.e.  guiding the robot through physical contact~\citep{zhu2018robotLearning}. Recent advance in deep learning and imitation learning has allowed the robot arm to learn to perform manipulation tasks by extracting knowledge from human demonstration within videos, eliminating the need of a human to teach the robot through physical contact, thus increasing efficiency and safety \citep{yu2018oneShot, yang2019learningActions}. In the future, we will further improve our one-shot teaching procedure by exploring state-of-the-art deep learning and imitation learning approaches. 
}

\textcolor{black}{
The colored point cloud registration used in the proposed IPE algorithm for recovering 6D parking pose is accurate but slow.  Furthermore, the cheap "Intel Realsense D435" depth sensor used in the paper is sensitive to the ambient lighting and temperature, which will have a negative impact on the retrieved point cloud's quality, resulting in bad point cloud registration results. In future work, a faster and more robust 6D pose estimation algorithm will be explored. Depth fusion algorithms (e.g.~\citep{sdf-man,UDFNet}) will be explored using a cheap depth sensor (e.g.: Intel Realsense D435) rather than the expensive sensors (e.g.: Pickit 3D, Zivid) to improve the input point cloud's quality as well. Additionally, the proposed framework in this paper will be further tested in more realistic environments that accurately resemble real-world scenarios such as factory manufacturing, food servicing, quality inspection, etc..  
}

\textcolor{black}{
Obstacle avoidance is necessary for the robot arm to work in unpredictable dynamic environments where safety is crucial, such as environments with constant human-robot interaction\citep{lin2017realTime}. The work in this paper focuses on industrial environments such as automated factories. Considering that there is little human interaction within automated factories, the environment remains unchanged throughout the operation. As long as the trajectory is planned and executed properly, it is unlikely for collisions to happen. Thus, collision avoidance was not necessary and therefore not the focus of this work. Additionally, the robot arm has collision detection functionality built-in within the controller as a safety mechanism by the robot arm manufacturer. Whenever the robot arm senses a collision with an object, it will perform an emergency stop to prevent damage \citep{haddadin2017robotCollisions}. For future work, collision avoidance can be implemented to allow the mobile manipulator to operate in environments that have uncertain changes over time due to human interactions, such as restaurants. Further hardware upgrades will be done to include more depth cameras, as the current depth camera mounted on the robot arm is not able to construct the real-time $360^{\circ}$ 3D map of the surroundings needed for collision avoidance algorithms. 
}

\textcolor{black}{
	A proper physical gripper is important for a successful manipulation of the rigid or non-rigid items. In the real-life setting, the engineers have to design the proper grippers for different kinds of items which need to be manipulated. The manual design process is time-consuming and expensive. In future work, we will explore the possibility of developing algorithms to autonomously design the physical grippers given the 3D model of the target items. 
}

\section{Conclusions}\label{Conclusion}
This paper presents a framework for flexible manufacturing that allows a robot to redo multiple industrial tasks at different sites in a hostile industrial environment, after one-shot teaching by a human operator. The framework consists of two stages: the teaching stage and the automation stage. During the teaching stage, the location of the robot base and the path of the robot arm's end-effector have been recorded. During automation stage, the aim is to make the robot arm's end-effector repeat a path as same as the reference path in the world frame. The key for mobile manipulator process automation is the accurate estimation of the relative 6D parking pose between the teaching stage and the automation stage, which is used to adjust the path of the robot arm end-effector. Thus, we propose the IPE algorithm to estimate an accurate relative 6D pose by registering to the camera's initial sampling pose (at teaching stage) in the world coordinate system iteratively.

\printcredits

\section*{Declaration of Competing Interest}
The authors declare that they have no known competing financial interests or personal relationships that could have appeared to influence the work reported in this paper.

\section*{Acknowledgement}
The research funding is from Shenzhen Amigaga Technology Co. Ltd. by the Gagabot2022 project (Grant Agreement No. P987001), from the Human Resources and Social Security Administration of Shenzhen Municipality by Overseas High-Caliber Personnel project (Grant NO. 202102222X, Grant NO. 202107124X) and from Human Resources Bureau of Shenzhen Baoan District by High-Level Talents in Shenzhen Baoan project (Grant No. 20210400X, Grant No. 20210402X) .

\appendix
\section*{Appendix A. Ablation Study}
\renewcommand\thefigure{\Alph{section}.\arabic{figure}} 
\renewcommand\thetable{\Alph{section}.\arabic{table}}
\refstepcounter{section}
\setcounter{figure}{0}    
\setcounter{table}{0}

\textcolor{black}{
In this part, we will test the impact on the IPE's performance from the pose difference threshold $\boldsymbol{\alpha}$, the IPE's maximum iteration number $\beta$ and the point clouds from the scenes used in IPE. In each trial, the experiment setting is the same as that in Section~\ref{IPE-performance} except for the following factors that are to be compared: the pose difference threshold,  the IPE's maximum iteration number, and the point cloud for registration from the scene. Equation~\eqref{eq:23} and Equation~\eqref{eq:24} are used to evaluate the rotation and translation error briefly rather than the mean and standard deviation of the 6D pose vector $[tx, ty, tz, r, p, y]$ (short for [translation on $x$ axis, translation on $y$ axis, translation on $z$ axis, roll, pitch, yaw]). We define that successful trial as a trial whose final pose difference is below the pose difference threshold $\boldsymbol{\alpha}$ and whose total iteration number does not exceed the IPE maximum iteration number $\beta$. We use the success rate $suc$ to represent the rate between the number of the successful trials (which have converged successfully) and the total number of the trials. The average iteration number $\overline{L}$ represents the average iteration number required for the successful trials. $\overline{E_R}$ and $\sigma_{E_R}$ are the mean and standard deviation of the rotation error only considering the successful trials. $\overline{E_t}$ and $\sigma_{E_t}$ are the mean and standard deviation of the translation error (unit: meter) only considering the successful trials.    
}

\subsection{\textcolor{black}{The pose difference threshold $\boldsymbol{\alpha}$}}
\textcolor{black}{
 In this module, the threshold $\boldsymbol{\alpha}$ of the pose difference $(x, y, z, roll, pitch, yaw)$ will be set as $\boldsymbol{\alpha_1}$ = (0.001m, 0.001m, 0.001m, $0.25^\circ$, $0.25^\circ$, $0.25^\circ$), $\boldsymbol{\alpha_2}$ = (0.0015m, 0.0015m, 0.0015m, $0.375^\circ$, $0.375^\circ$, $0.375^\circ$), $\boldsymbol{\alpha_3}$ = (0.002m, 0.002m, 0.002m, $0.5^\circ$, $0.5^\circ$, $0.5^\circ$),  $\boldsymbol{\alpha_4}$ = (0.0025m, 0.0025m, 0.0025m, $0.625^\circ$, $0.625^\circ$, $0.625^\circ$), $\boldsymbol{\alpha_5}$ = (0.003m, 0.003m, 0.003m, $0.75^\circ$, $0.75^\circ$, $0.75^\circ$). The $\boldsymbol{\alpha}$ original setting is $\boldsymbol{\alpha_3}$ = (0.002m, 0.002m, 0.002m, $0.5^\circ$, $0.5^\circ$, $0.5^\circ$) in Section~\ref{IPE-performance}. On each site, we have done 30 trials for each  $\boldsymbol{\alpha}$ setting separately.  Table~\ref{table:alpha-ablation-study-site1}, Table~\ref{table:alpha-ablation-study-site2} and Table~\ref{table:alpha-ablation-study-site3} show how $\boldsymbol{\alpha}$ affects the IPE performance.  With $\boldsymbol{\alpha}$ increasing, the success rate $suc$ will increase and the average iteration number $\overline{L}$ will decrease. The IPE accuracy increases when $\boldsymbol{\alpha}$ decreases. Considering the tradeoff among the success rate, average iteration number and accuracy, the optimal value should be $\boldsymbol{\alpha_3}$ = (0.002m, 0.002m, 0.002m, $0.5^\circ$, $0.5^\circ$, $0.5^\circ$).   
}

\textcolor{black}{
	\begin{table}[h!]
		\centering
		\caption{The factor $\boldsymbol{\alpha}$ ablation study for Site 1} \label{table:alpha-ablation-study-site1}
		\begin{tabular}{|p{0.75cm}|| p{1cm}| p{1cm}| p{1cm}| p{1cm}| p{1cm}|}
			\hline 
			\textit{Factor}  &  \textit{$\boldsymbol{\alpha_1}$} & \textit{$\boldsymbol{\alpha_2}$}  & \textit{$\boldsymbol{\alpha_3}$} & \textit{$\boldsymbol{\alpha_4}$}  & \textit{$\boldsymbol{\alpha_5}$} \\
			\hline 
			$suc$ &   0.8333 & 0.9667 & 1.0000 & 1.0000 & 1.0000 \\
			\hline 
			$\overline{L}$ & 4.0800  & 3.8276 & 3.5667 & 3.2667 & 2.9667 \\ 
			\hline 
			$\overline{E_R}$ & 0.0168 & 0.0168 & 0.0169 & 0.0199 & 0.0215 \\ 
			\hline 
			$\sigma_{E_R}$ &   0.0074 & 0.0075 & 0.0074 & 0.0104 & 0.0115 \\
			\hline 
			$\overline{E_t}$ / (m) & 0.0048 & 0.0049 & 0.0049 & 0.0050 & 0.0052 \\ 
			\hline 
			$\sigma_{{E_t}}$ / (m) & 0.0024 & 0.0025 & 0.0025 & 0.0026 & 0.0027 \\ 
			\hline		
		\end{tabular} 	
	\end{table}   
}

\textcolor{black}{
	\begin{table}[h!]
		\centering
		\caption{The factor $\boldsymbol{\alpha}$ ablation study for Site 2} \label{table:alpha-ablation-study-site2}
		\begin{tabular}{|p{0.75cm}|| p{1cm}| p{1cm}| p{1cm}| p{1cm}| p{1cm}|}
			\hline 
			\textit{Factor}  &  \textit{$\boldsymbol{\alpha_1}$} & \textit{$\boldsymbol{\alpha_2}$}  & \textit{$\boldsymbol{\alpha_3}$} & \textit{$\boldsymbol{\alpha_4}$}  & \textit{$\boldsymbol{\alpha_5}$} \\
			\hline 
			$suc$ &   0.8667 & 0.9667 & 1.0000 & 1.0000 & 1.0000 \\
			\hline 
			$\overline{L}$ & 3.5385  & 2.9310 & 2.8000 & 2.5000 & 2.3667 \\ 
			\hline 
			$\overline{E_R}$ & 0.0175 & 0.0177 & 0.0183 & 0.0185 & 0.0186 \\ 
			\hline 
			$\sigma_{E_R}$ &   0.0063 & 0.0064 & 0.0063 & 0.0065 & 0.0065 \\
			\hline 
			$\overline{E_t}$ / (m) & 0.0054 & 0.0054 & 0.0054 & 0.0055 & 0.0055 \\ 
			\hline 
			$\sigma_{{E_t}}$ / (m) & 0.0018 & 0.0019 & 0.0019 & 0.0020 & 0.0022 \\ 
			\hline		
		\end{tabular} 	
	\end{table}   
}

\textcolor{black}{
	\begin{table}[h!]
		\centering
		\caption{The factor $\boldsymbol{\alpha}$ ablation study for Site 2'} \label{table:alpha-ablation-study-site3}
		\begin{tabular}{|p{0.75cm}|| p{1cm}| p{1cm}| p{1cm}| p{1cm}| p{1cm}|}
			\hline 
			\textit{Factor}  &  \textit{$\boldsymbol{\alpha_1}$} & \textit{$\boldsymbol{\alpha_2}$}  & \textit{$\boldsymbol{\alpha_3}$} & \textit{$\boldsymbol{\alpha_4}$}  & \textit{$\boldsymbol{\alpha_5}$} \\
			\hline 
			$suc$ &   0.9667 & 0.9667 & 1.0000 & 1.0000 & 1.0000 \\
			\hline 
			$\overline{L}$ & 5.8966  & 5.1379 & 4.8000 & 4.1667 & 4.0000 \\ 
			\hline 
			$\overline{E_R}$ & 0.0224 & 0.0224 & 0.0225 & 0.0227 & 0.0228 \\ 
			\hline 
			$\sigma_{E_R}$ &   0.0102 & 0.0103 & 0.0103 & 0.0105 & 0.0106 \\
			\hline 
			$\overline{E_t}$ / (m) & 0.0088 & 0.0088 & 0.0089 & 0.0089 & 0.0090 \\ 
			\hline 
			$\sigma_{{E_t}}$ / (m) & 0.0042 & 0.0044 & 0.0044 & 0.0045 & 0.0045 \\ 
			\hline		
		\end{tabular} 	
	\end{table}   
}

\subsection{\textcolor{black}{IPE maximum iteration number $\beta$}}
\textcolor{black}{
	In this module, the threshold $\boldsymbol{\beta}$ of IPE's maximum iteration number will be set as $\boldsymbol{\beta_1}$ = 3, $\boldsymbol{\beta_2}$ = 4, $\boldsymbol{\beta_3}$ = 5, $\boldsymbol{\beta_4}$ = 6, $\boldsymbol{\beta_5}$ = 7 for site 1 and site 2. The threshold $\boldsymbol{\beta}$ of IPE's maximum iteration number will be set as $\boldsymbol{\beta_1}$ = 8, $\boldsymbol{\beta_2}$ = 9, $\boldsymbol{\beta_3}$ = 10, $\boldsymbol{\beta_4}$ = 11, $\boldsymbol{\beta_5}$ = 12 for site 2'. The $\beta$ original setting is 5  for site 1, site 2 and 10 for site 2' in Section~\ref{IPE-performance} , which is equal to $\boldsymbol{\beta_3}$ on the corresponding site. On each site, we have done 30 trials for each  $\beta$ setting separately.  Table~\ref{table:beta-ablation-study-site1}, Table~\ref{table:beta-ablation-study-site2} and Table~\ref{table:beta-ablation-study-site3} show how $\beta$ affects the IPE performance. With the increase of $\beta$, the success rate $suc$ will increase. The average iteration number $\overline{L}$ will increase drastically from $\beta_1$ to $\beta_3$ and stay steady from $\beta_3$ to $\beta_5$ (Note: $suc = 1$ for $\beta_3$, $\beta_4$ and $\beta_5$). IPE's accuracy stays nearly the same from $\beta_1$ to $\beta_5$ because they use the same $\boldsymbol{\alpha}$ setting and $\boldsymbol{\alpha}$ controls the convergence accuracy. Considering the tradeoff among the success rate, average iteration number and accuracy, the optimal value should be $\beta = \beta_3$. It should be noted that increasing $\beta$ does not guarantee correct convergence. There may exist a case where the IPE is unable to converge correctly (e.g.: in the scene - a plane of pure color), increasing $\beta$ in such case will only waste time.}

\textcolor{black}{
	\begin{table}[h!]
		\centering
		\caption{The factor $\beta$ ablation study for Site 1} \label{table:beta-ablation-study-site1}
		\begin{tabular}{|p{0.75cm}|| p{1cm}| p{1cm}| p{1cm}| p{1cm}| p{1cm}|}
			\hline 
			\textit{Factor}  &  \textit{$\beta_1$} & \textit{$\beta_2$}  & \textit{$\beta_3$} & \textit{$\beta_4$}  & \textit{$\beta_5$} \\
			\hline 
			$suc$ &   0.9000 & 0.9000 & 1.0000 & 1.0000 & 1.0000 \\
			\hline 
			$\overline{L}$ & 2.8148  & 3.2593 & 3.5667 & 3.5333 & 3.6000 \\ 
			\hline 
			$\overline{E_R}$ & 0.0172 & 0.0170 & 0.0169 & 0.0168 & 0.0169 \\ 
			\hline 
			$\sigma_{E_R}$ &   0.0080 & 0.0073 & 0.0074 & 0.0074 & 0.0073 \\
			\hline 
			$\overline{E_t}$ / (m) & 0.0052 & 0.0050 & 0.0049 & 0.0048 & 0.0048 \\ 
			\hline 
			$\sigma_{{E_t}}$ / (m) & 0.0029 & 0.0025 & 0.0025 & 0.0024 & 0.0023 \\ 
			\hline		
		\end{tabular} 	
	\end{table}   
}

\textcolor{black}{
	\begin{table}[h!]
		\centering
		\caption{The factor $\beta$ ablation study for Site 2} \label{table:beta-ablation-study-site2}
		\begin{tabular}{|p{0.75cm}|| p{1cm}| p{1cm}| p{1cm}| p{1cm}| p{1cm}|}
			\hline 
			\textit{Factor}  &  \textit{$\beta_1$} & \textit{$\beta_2$}  & \textit{$\beta_3$} & \textit{$\beta_4$}  & \textit{$\beta_5$} \\
			\hline 
			$suc$ &   0.8667 & 0.9333 & 1.0000 & 1.0000 & 1.0000 \\
			\hline 
			$\overline{L}$ & 2.5385  & 2.6429 & 2.8000 & 2.8333 & 2.8000 \\ 
			\hline 
			$\overline{E_R}$ & 0.0184 & 0.0183 & 0.0183 & 0.0183 & 0.0183 \\ 
			\hline 
			$\sigma_{E_R}$ &   0.0065 & 0.0063 & 0.0063 & 0.0064 & 0.0063 \\
			\hline 
			$\overline{E_t}$ / (m) & 0.0056 & 0.0054 & 0.0054 & 0.0054 & 0.0055 \\ 
			\hline 
			$\sigma_{{E_t}}$ / (m) & 0.0021 & 0.0020 & 0.0019 & 0.0018 & 0.0018 \\ 
			\hline		
		\end{tabular} 	
	\end{table}   
}

\textcolor{black}{
	\begin{table}[h!]
		\centering
		\caption{The factor $\beta$ ablation study for Site 2'} \label{table:beta-ablation-study-site3}
		\begin{tabular}{|p{0.75cm}|| p{1cm}| p{1cm}| p{1cm}| p{1cm}| p{1cm}|}
			\hline 
			\textit{Factor}  &  \textit{$\beta_1$} & \textit{$\beta_2$}  & \textit{$\beta_3$} & \textit{$\beta_4$}  & \textit{$\beta_5$} \\
			\hline 
			$suc$ &   0.9333 & 0.9667 & 1.0000 & 1.0000 & 1.0000 \\
			\hline 
			$\overline{L}$ & 4.3929  & 4.6207 & 4.8000 & 4.8333 & 4.8333 \\ 
			\hline 
			$\overline{E_R}$ & 0.0226 & 0.0225 & 0.0225 & 0.0225 & 0.0225 \\ 
			\hline 
			$\sigma_{E_R}$ &   0.0104 & 0.0104 & 0.0103 & 0.0102 & 0.0103 \\
			\hline 
			$\overline{E_t}$ / (m) & 0.0091 & 0.0090 & 0.0089 & 0.0089 & 0.0088 \\ 
			\hline 
			$\sigma_{{E_t}}$ / (m) & 0.0046 & 0.0044 & 0.0044 & 0.0044 & 0.0044 \\ 
			\hline		
		\end{tabular} 	
	\end{table}   
}

\subsection{\textcolor{black}{The point clouds from the scenes}}
\begin{figure*}
	\begin{center}
		\subfloat{\includegraphics[width=1\linewidth]{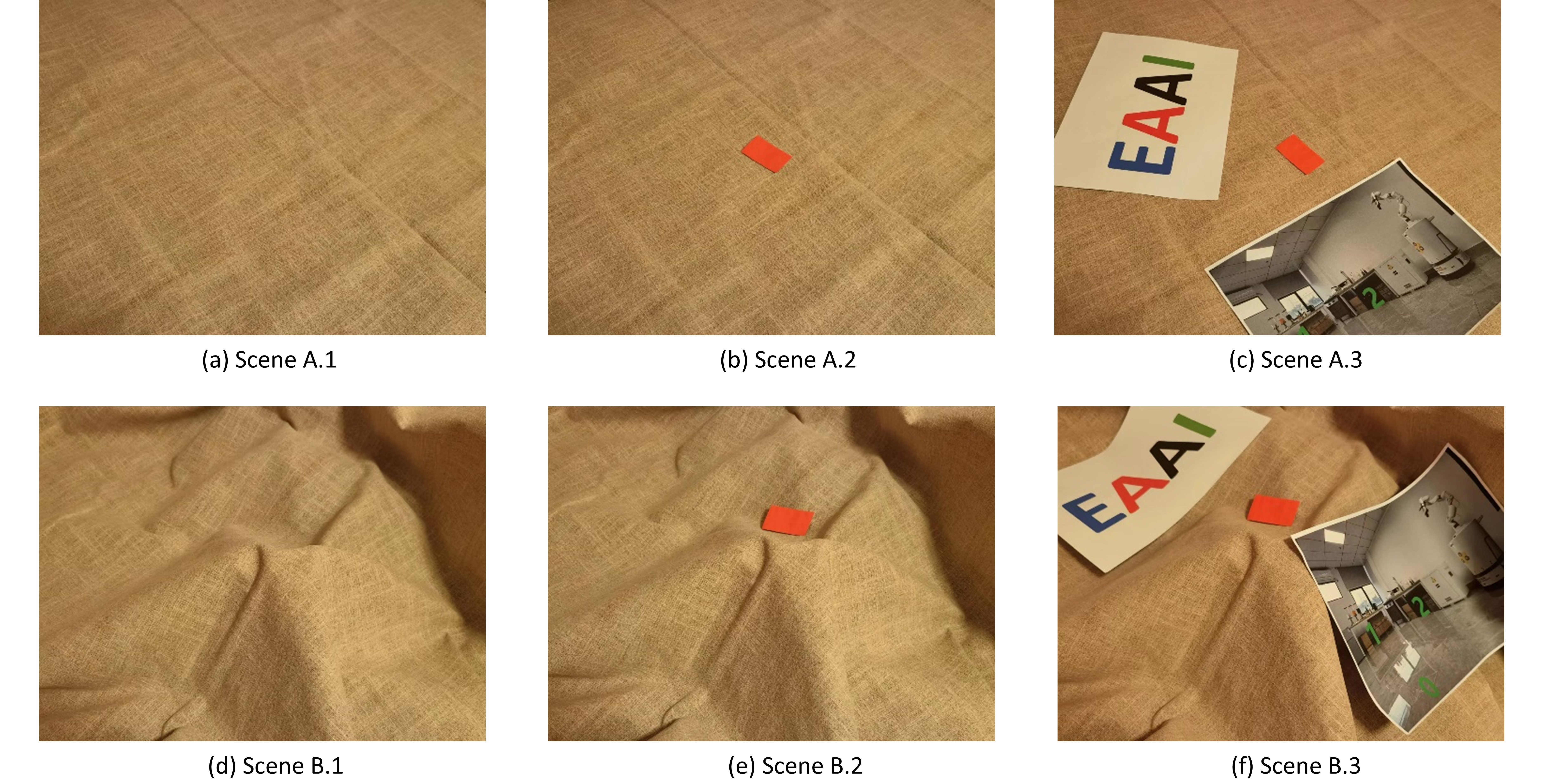}}	
	\end{center}
	\caption{More Scenes for Test.}
	\label{fig:point-cloud-ablation-study-appendix}
\end{figure*}

\textcolor{black}{
	In this module, more scenes have been used to test the robustness and accuracy of IPE by considering the geometry and color of the scene. In Figure~\ref{fig:point-cloud-ablation-study-appendix}: (a) Scene A.1 shows the scene containing a single plane with single pure color; (b) Scene A.2 shows a scene based on the Scene A.1 with a minor variation of the color using a red dot (which is printed on a thin paper); (c) Scene A.3 shows a scene based on Scene A.1 with rich color and textures (which are printed on three piece of thin paper); (d) Scene B.1 shows the scene with more geometry features but still with pure color; (e) Scene B.2 shows a scene based on Scene B.1 with  a minor variation of the color by using a red dot (which is printed on a thin paper); (f) Scene B.3 shows a scene based on Scene B.1 with rich color using much textures (which are printed on three piece of thin paper); }. 

\textcolor{black}{
There are 6 groups of experiments where each group of experiments corresponds to a scene. Experiment group Ex A.1, Ex A.2, and Ex A.3 corresponds to Scene A.1, A.2, A.3 respectively. Experiment group Ex B.1, Ex B.2, and Ex B.3 corresponds to Scene B.1, B.2, B.3 respectively. There are 2 additional groups of experiments where the color of the scene is not used in IPE\footnote{Given the input point cloud has no color and the colored point cloud registration algorithm~\citep{colored-point-cloud-registration-2017} is not capable of dealing with it, we replace the colored point cloud registration algorithm used in IPE with ICP~\citep{ICP1992}.}. We remove the color information of the point cloud from Scene A.1 and Scene B.1, and treat the corresponding data in Experiment group Ex A.0 and Ex B.0 respectively. Each group of experiments consists of 10 trials. $\boldsymbol{\alpha}$ = (0.002m, 0.002m, 0.002m, $0.5^\circ$, $0.5^\circ$, $0.5^\circ$), $\beta$ =5. Table~\ref{table:scene-ablation-study-site1} lists the performance result of IPE with different point cloud input from various scenes. Ex A.0 - Ex A.2 shows the performance of IPE will collapse when the geometry feature is too poor and there is no or little colorful texture. Ex A.3 shows IPE could work when there is rich colorful texture but poor geometry feature. The reason why the success rate is low in Ex A.3 is that the global point cloud registration~\citep{fast-global-registration2016} in IPE breaks down when the geometry features are poor. If the global registration algorithm could not provide an initial transform to the local colored point cloud registration algorithm~\citep{colored-point-cloud-registration-2017}, the colored point cloud registration algorithm is easy to fail when the initial transformation between the two registered point clouds is big. Ex B.0 - Ex B.3 shows rich geometry features contribute to a higher success rate. The rich colorful texture contributes to a higher precision.   
}

\textcolor{black}{
	\begin{table*}[h!]
		\centering
		\caption{\textcolor{black}{IPE performance on different scenes}} \label{table:scene-ablation-study-site1}
		\begin{tabular}{|p{1.5cm}|| p{1.4cm}| p{1.4cm}| p{1.4cm}| p{1.4cm}| p{1.4cm}|  p{1.4cm}| p{1.4cm}| p{1.4cm}|}
			\hline 
			\textit{Factor}  &  Ex A.0  & Ex A.1  & Ex A.2 & Ex A.3  & Ex B.0 & Ex B.1  & Ex B.2 & Ex B.3 \\
			\hline 
			$suc$ &   0 & 0 & 0 & 0.5000 & 0.8000  & 0.9000 & 0.9000 & \bf{1.0000} \\
			\hline 
			$\overline{L}$ & Null  & Null & Null & 3.6000 & 3.7500 & \bf{3.5556} & 3.6667 & 3.6000 \\ 
			\hline 
			$\overline{E_R}$ & Null & Null & Null & 0.0160 & 0.0217 & 0.0204 & 0.0206 & \bf{0.0155} \\ 
			\hline 
			$\sigma_{E_R}$ &   Null & Null & Null & 0.0078 & 0.0118 & 0.0089 & 0.0073 & \bf{0.0072} \\
			\hline 
			$\overline{E_t}$ / (m) & Null & Null & Null & 0.0047 & 0.0048 & 0.0049 & 0.0047 & \bf{0.0046} \\ 
			\hline 
			$\sigma_{{E_t}}$ / (m) & Null & Null & Null & \bf{0.0023} & 0.0025 & 0.0026 & 0.0025 & \bf{0.0023} \\ 
			\hline		
		\end{tabular} 	
	\end{table*}   
}

\refstepcounter{section}
\section*{\textcolor{black}{Appendix B. List of Abbreviations}} \label{Abbreviations}
\noindent 
\textcolor{black}{
	\begin{tabular}{@{}ll}
		MMPA & Mobile Manipulator Process Automation \\
		& with One-shot Teaching\\
		SLAM & Simultaneous Localization And Mapping\\
		IPE & Iterative Pose Estimation by Eye \& Hand\\
		ICP & Iterative Closest Point \\
		IMU & Inertial Measurement Unit \\
		QR & Quick Response \\
		DOF & Degree of Freedom \\
		RRT* & Rapidly Exploring Random Tree \\
		PRM & Probabilistic Roadmap \\			
\end{tabular}}

\refstepcounter{section}
\section*{\textcolor{black}{Appendix C. List of Symbols}} \label{symbols}
\noindent 
\textcolor{black}{
	\begin{tabular}{@{}ll}
		$XYZO_{MB}$ & Mobile Base Coordinate System\\
		$XYZO_{B}$ & Robot Arm Base Coordinate System\\
		$XYZO_{C}$ & Depth Camera Coordinate System\\
		$XYZO_{W}$ & World Coordinate System\\
		$XYZO_{n}$& Coordinate System for $n^{th}$ Robot Joint\\
		$tk_{i}$ & Task $i$\\
		$L_{i}(vehicle) $ & Mobile Platform Parking location for \\
		& Task $i$ in Teaching Stage\\
		$X_{i}(O)$ & Colored 3D Point Cloud for Task $i$ \\
		&in Teaching Stage\\
		$Path_{i}(O)$ & Robot Arm Path Information for Task $i$ \\
		&in Teaching Stage\\
		$\tilde{L}_{i}(vehicle)$ & Mobile Platform Parking location for \\
		& Task $i$ in Automation Stage\\			
		$X_{i}(k)$ & Colored 3D Point Cloud for Task $i$ \\
		&in  $k^{th}$ Iteration in Automation Stage\\
		$\widetilde{Path}_{i}(arm)$ & Adapted Robot Arm Path for Task $i$\\
		$\Delta R_{C_{k}}$, $\Delta t_{C_{k}}$ & Camera's Relative Rotation and \\
		& Translation Matrix between Teaching \\
		& Stage and $k^{th}$ Iteration in Automation Stage\\
		$\Delta R_{B_{k}}$, $\Delta t_{B_{k}}$ & Robot Arm's Estimated Relative Rotation \\
		& and Translation Matrix between Teaching \\
		&Stage and $k^{th}$ Iteration in Automation Stage\\
		$R^{C_o}_{B_o}$, $t^{C_o}_{B_o}$ & Rotation and Translation Matrix from \\
		& Coordinate $XYZO_{C}$ to $XYZO_{B}$ \\
		&in Teaching Stage\\
		$R^{C_o}_{B_k}$, $t^{C_k}_{B_k}$ & Rotation and Translation Matrix from \\
		&Coordinate $XYZO_{C}$ to $XYZO_{B}$ \\
		&after $k^{th}$ Iteration Automation Stage\\
		$S_{C_o}$ & One 3D Point in Coordinate $XYZO_{C}$ in \\
		&Teaching Stage\\
		$S_{B_o}$ & One 3D Point in Coordinate $XYZO_{B}$ in \\
		&Teaching Stage\\
		$S_{C_k}$ & One 3D Point in Coordinate $XYZO_{C}$ in \\
		&$k^{th}$ Iteration Automation Stage\\
		$S_{B_k}$ & One 3D Point in Coordinate $XYZO_{B}$ in \\
		&$k^{th}$ Iteration in Automation Stage\\
		$\boldsymbol{t_{est}, R_{est}}$ & Estimated Translation and Rotation Matrix \\
		$\boldsymbol{t_{gt}, R_{gt}}$ & Ground Truth Translation and Rotation \\
		& Matrix \\
		$E_{R}$ & Rotation Error \\
		$E_{t}$ & Translation Error \\
		$\alpha_{n}$ & Pose Difference Threshold for Test Case $n$ \\
		$\beta_{n}$ & Pose Difference Threshold for Test Case $n$ \\
		$suc$ & Success rate \\
		$\overline{L}$ & Average Iteration Number Required for Success\\
		$\overline{E_R}$, $\sigma_{E_R}$ & Mean and Standard Deviation of the Rotation \\
		& Error Considering Only Successful Trials\\
		$\overline{E_t}$, $\sigma_{E_t}$ & Mean and Standard Deviation of the Translation \\
		& Error Considering Only Successful Trials\\
\end{tabular}}

\bibliographystyle{cas-model2-names}

\bibliography{cas-refs}

\end{document}